\documentclass{article}

\usepackage{microtype}
\usepackage{graphicx}
\usepackage{subcaption}
\usepackage{booktabs} % for professional tables

\usepackage[utf8]{inputenc} % allow utf-8 input
\usepackage[T1]{fontenc}    % use 8-bit T1 fonts
\usepackage{hyperref}       % hyperlinks
\usepackage{url}            % simple URL typesetting
\usepackage{amsfonts}       % blackboard math symbols
\usepackage{nicefrac}       % compact symbols for 1/2, etc.
\usepackage{xcolor}         % colors
\usepackage{caption} 

\usepackage{colortbl}
\usepackage{algorithm}
\usepackage{amsmath}
\usepackage{algpseudocode}
\usepackage{algorithmicx}
\usepackage{algcompatible}
\usepackage{cancel}
\usepackage{rotating}
\usepackage{placeins} 

\usepackage{wrapfig}
\usepackage{graphicx}
\usepackage{enumitem}
\usepackage{multirow}

\usepackage{adjustbox}

\usepackage{subcaption}
\usepackage[table]{xcolor}

\usepackage[preprint]{icml2026}

\usepackage{amsmath}
\usepackage{amssymb}
\usepackage{mathtools}
\usepackage{amsthm}

\usepackage[capitalize,noabbrev]{cleveref}

\theoremstyle{plain}

\theoremstyle{definition}

\theoremstyle{remark}

\usepackage[textsize=tiny]{todonotes}

\icmltitlerunning{Contrastive Residual Energy Test-time Adaptation}

\begin{document}

\twocolumn[
  \icmltitle{Contrastive Residual Energy Test-time Adaptation}



\icmlsetsymbol{equal}{*}

  \begin{icmlauthorlist}
    \icmlauthor{Yewon Han}{equal,yyy}
    \icmlauthor{Seoyun Yang}{equal,yyy}
    \icmlauthor{Taesup Kim}{yyy}
  \end{icmlauthorlist}

  \icmlaffiliation{yyy}{Graduate School of Data Science, Seoul National University, Seoul, Korea}

  \icmlcorrespondingauthor{Taesup Kim}{taesup.kim@snu.ac.kr}

  \icmlkeywords{Machine Learning, ICML}

  \vskip 0.3in
]


% is text to display at the start of the footnote. The 

\printAffiliationsAndNotice{\icmlEqualContribution} % no special notice (required even if empty)

\begin{abstract}

Test-time adaptation (TTA) enhances model robustness by enabling adaptation to target distributions that differ from training distributions, improving real-world generalizability. 
However, most existing TTA approaches focus on adjusting the conditional distribution and therefore exhibit poor calibration, as they rely on uncertain predictions in the absence of labels. 
Energy-based TTA frameworks provide an alternative by modeling the marginal distribution of target data without depending on label predictions, but their reliance on costly sampling hinders scalability in real-world scenarios where decisions must be made without latency.
In this work, we propose \textit{Contrastive Residual Energy Test-time Adaptation (\textsc{CreTTA})}, a practical solution for reliable adaptation. 
We theoretically reformulate the marginal distribution adaptation as learning a residual energy function. 
This formulation leads to a contrastive objective where the intractable partition function mathematically cancels out, removing sampling and approximation error.
Crucially, our analysis reveals that this design prevents overfitting through an adaptive gradient reweighting mechanism that leverages relative energy differences, avoiding the self-confirming bias of entropy minimization. 
Extensive experiments demonstrate that \textsc{CreTTA} achieves scalable and well-calibrated adaptation under real-world computational constraints.

\end{abstract}    
\section{Introduction}
\label{sec:intro}
% about tta 
Deep learning models can achieve high accuracy on training and testing data from the same distribution.
However, when the distribution of the test data diverges from the original training dataset, the performance of the deep learning models deteriorates. 
This \textit{distribution shift} refers to changes in the underlying data statistics, such as feature distributions or environmental conditions, between training and deployment.
It is a major challenge in real-world scenarios, where test samples are often drawn from a distribution that deviates from the training data.

% ----- 
% \begin{figure}
% % \vspace{-20pt} 
% \centering
%  \includegraphics[width=0.90\linewidth]{figures/IMAGENETC_ECE_ACC.pdf}
% \caption{ImageNet-C (Sev 5) ECE($\downarrow$) and Acc($\uparrow$) over batch progress. \textsc{CreTTA} maintains stable calibration performance, while TEA experiences approximation error accumulation.}    
% \label{fig:imagenetc_ece_acc}
% \small
% % \vspace{-20pt}
% \vspace{-1.0\baselineskip}
% \end{figure}
% \index{figure}
% ----- 

\begin{figure}
\centering
 \includegraphics[width=1.0\linewidth]{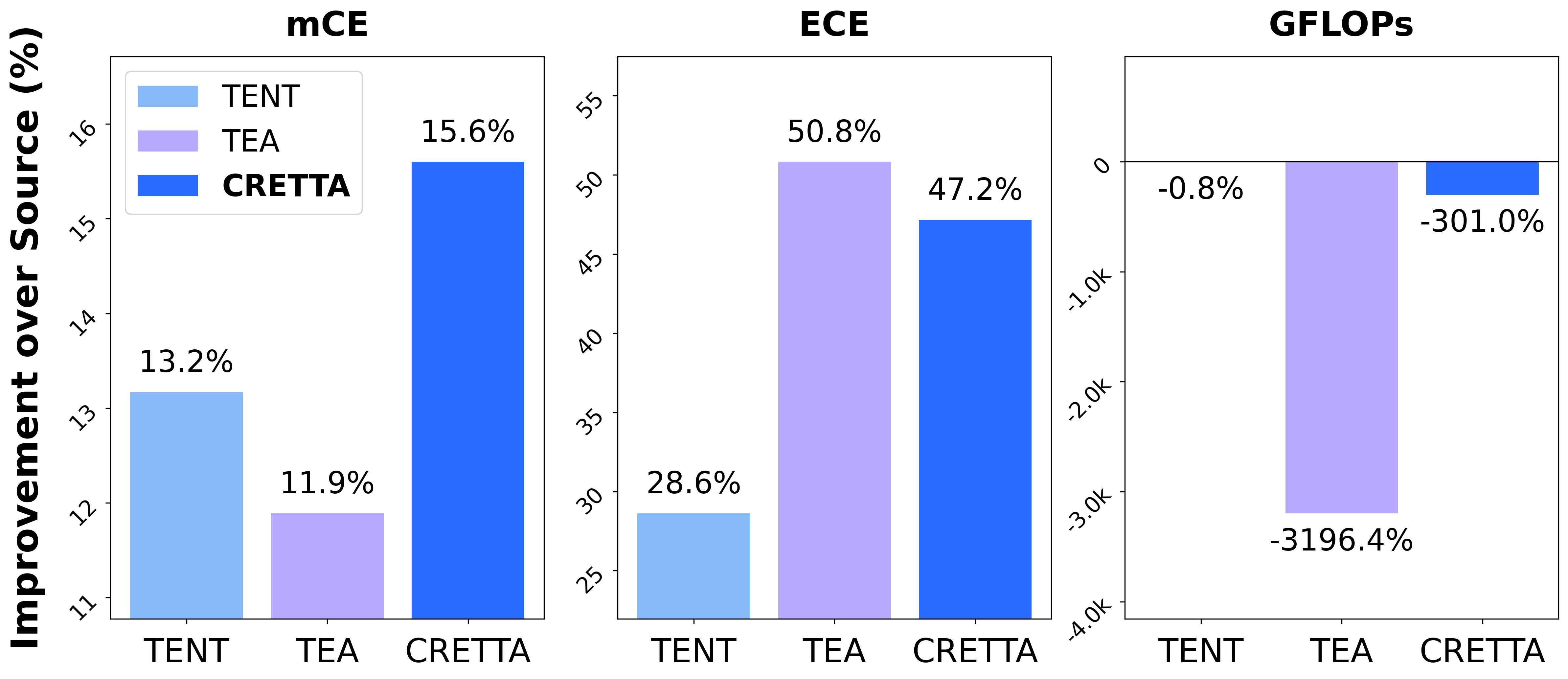}
 \caption{Comparison of classification (mCE), calibration (ECE) performance gains over source with corresponding computational costs (GFLOPs) to achieve improvements on CIFAR10-C (Sev 1-5 Avg). {\textsc{CreTTA} achieves a better balance between performance and computation} than the existing energy-based method, TEA.}
 
% \caption{Comparison of classification performance (mCE), calibration performance (ECE) and computational cost (GFLOPs) required to achieve such performance over source for baselines on CIFAR10-C
% Acc and ECE improvement over source for baselines on the CIFAR10-C}    

\label{fig:comcost}
\vspace{-\baselineskip}
\end{figure}
\index{figure}
%-------------------

To address distribution shifts during testing, \textit{test-time adaptation} (TTA) strategy aims to adapt trained model instantly, thereby maintaining robust performance on unexpected out-of-distribution samples. 
Since ground-truth labels are unavailable at test-time, existing approaches~\citep{lee2013pseudo, liang2021reallyneedaccesssource} often rely on the model's own predictions.
For instance, TENT~\citep{wang2021tentfullytesttimeadaptation} uses the model's predicted probability distribution as a surrogate ground-truth within an Entropy-Minimization(EM) objective.
Formally, the entropy minimization objective for a test sample $x_t$ is expressed as $-\sum_{k=1}^{C} p_{\theta}(\hat{y}_k \mid x_t) \log p_{\theta}(\hat{y}_k \mid x_t)$, where $C$ is the number of classes and $p_{\theta}(\hat{y}_k \mid x_t)$ is the predicted probability of class $\hat{y}_k$. 
While this approach demonstrates promising accuracy, it fundamentally relies on uncertain predictions without ground-truth supervision.
As a result, optimizing the entropy minimization objective creates a \textit{self-confirming bias}, driving the predicted probabilities to collapse to extreme values of 0 or 1, leading to severe overconfidence~\citep{press2024entropyenigmasuccessfailure}.
This behavior increases calibration error as illustrated in \autoref{fig:overconfidence}, which can be detrimental in high-stakes real-world scenarios. 

\textit{Energy-based models} (EBMs)~\citep{ebmlecun} offer a promising alternative by directly modeling the marginal distribution, thereby avoiding the pitfalls of confidence-based objectives.
TEA~\citep{yuan2024teatesttimeenergyadaptation} applies Maximum Likelihood Estimation (MLE) in the TTA setting to align the model with unseen target distributions.
While providing experimental evidence of stable and reliable adaptation, standard EBM approaches still face two critical limitations:
% \vspace{-5pt}
\begin{itemize}[leftmargin=1em, itemsep=0.2em, topsep=0.3em] 
\item \textbf{Unresolvable Approximation Error.} The normalization constant requires approximation via short-run Markov chains, which yield \textit{biased gradient estimates}. Consequently, parameter updates can be unstable or converge to suboptimal solutions~\citep{song2021trainenergybasedmodels, yair2021contrastivedivergencelearningtime}, particularly during few-shot adaptation or with high-dimensional data as shown in \autoref{fig:imagenetc_ece_acc}.
\item \textbf{High Computational Cost.} Re-estimating the normalization constant for every incoming batch necessitates repeated sampling steps, incurring substantial computational overhead. This burden makes real-time deployment infeasible under \textit{strict latency constraints}, highlighting the need for methods that reduce overhead while maintaining strong performance.
\end{itemize}

Consequently, TTA faces a dilemma: existing methods are either (i) efficient but unreliable (entropy-based), or (ii) reliable but computationally prohibitive (sampling-based). 
In real-world settings, however, well-calibrated adaptation is critical to \textit{avoid overconfident errors while ensuring computationally efficient and scalable model behavior}.

To this end, we propose \textit{Contrastive Residual Energy Test-time Adaptation}~(\textsc{CreTTA}), a novel framework that optimizes the marginal distribution while eliminating the normalization constant approximation. 
We theoretically reformulate the TTA task as learning a \textit{residual energy function} that captures the discrepancy between source and target distributions.
By embedding this function into a pairwise \textit{contrastive objective}, we mathematically offset the intractable partition function, removing sampling and approximation error.

The primary contributions of our work are summarized as follows:
% \vspace{-5pt}
\begin{itemize}[leftmargin=1em, itemsep=0.2em, topsep=0.3em] 
    \item \textbf{Sampling-free EBM Framework.} We introduce \textsc{CreTTA}, a novel sampling-free residual energy-based TTA framework. By removing the sampling bottleneck, it offers significant \textit{computational efficiency} and \textit{scalability}, making it a practical solution for real-time applications.
    
    \item \textbf{Theoretical Reformulation and Stability.} We redefine the marginal distribution adaptation as learning a residual energy function with a contrastive objective. Our analysis reveals that this formulation acts as an implicit regularizer via an \textit{adaptive gradient reweighting mechanism}, which effectively mitigates overfitting and ensures stable adaptation.
    
    \item \textbf{Strong Performance with Reliability.} Extensive experiments demonstrate that \textsc{CreTTA} is well-calibrated and achieves state-of-the-art classification performance across various distribution shifts.
    % state-of-the-art performance across various distribution shifts, bridging the gap between reliability and accuracy.
\end{itemize}

\section{Preliminaries}
\label{sec:preliminaries}
%-------------------------------------------------------------------------
\paragraph{Problem Setup.}

% problem
Let $Q$ denote the marginal distribution of the source training data $x_s$. 
Consider a classifier $f_{\phi}(x)$ with parameters $\phi$, which is pretrained on a labeled source dataset $\{(x^{(i)}_s, y^{(i)}_s)\}_{i=1}^M$.
Although this pretrained model performs well on in-distribution test data (i.e., $x_s \sim Q$), its performance can degrade substantially when tested on data from a different distribution $P  (\neq Q)$, commonly referred to as out-of-distribution data.

% task & tent limitation (why MLE)
Test-time adaptation (TTA) aims to mitigate this issue by adapting the pretrained parameters $\phi$ to better align with the target marginal distribution $P$.
In this setting, we are given a set of $N$ unlabeled target samples $\{x_t^{(i)}\}_{i=1}^{N}$ drawn from $P$, which arrive in online batches.
To cope with the absence of label information, existing methods often rely on unsupervised objectives, particularly entropy minimization. 
EATA \citep{niu2022efficienttesttimemodeladaptation} and SAR \citep{niu2023stabletesttimeadaptationdynamic} build on this foundation by incorporating surrogate objectives and sample selection mechanisms to filter out unreliable predictions. 
However, these enhancements still fundamentally depend on uncertain model outputs, as dictated by the nature of entropy-based objectives.

Moving beyond entropy-based methods, recent work by TEA \citep{yuan2024teatesttimeenergyadaptation} demonstrates the promise of energy-based modeling for TTA, where the central idea is to represent the test data marginal distribution through an energy function.
Within this framework, TEA employs MLE on the marginal distribution of test samples {\scriptsize$\{x_t^{(i)}\}_{i=1}^{N}$}, so that the energy function is learned to assign lower energy (i.e., higher likelihood) to observed test inputs.
By directly modeling and adapting to the marginal distribution, TEA mitigates distribution shifts without requiring labeled data or relying heavily on potentially unreliable model predictions.
Building on this insight, our proposed method, \textsc{CreTTA}, approaches test-time adaptation not merely as an MLE procedure, but as effective learning of an energy function that represents an unnormalized marginal distribution.

%-------------------------------------------------------------------------
\paragraph{Energy-based Models.}

\citet{ebmlecun} express the marginal distribution in the energy-based model framework using Gibbs distribution, which can be formulated as 
$q_{\phi}(x) = \text{exp}(-E_{\phi}(x))/Z({\phi})$,
where $\phi \in \Phi$, with $\Phi$ representing the parameter space and $Z(\phi) = \int_x \exp(-E_\phi(x)) dx$ is the normalizing constant (partition function).
The energy function $E_{\phi}(x):\mathbb{R}^d \rightarrow\mathbb{R}$ maps a $d$-dimensional data point to a scalar energy value, thereby defining an unnormalized density over the data space.
The fundamental principle of EBMs is to represent the likelihood of a sample through this energy landscape: lower energy corresponds to higher likelihood and vice versa.
A well-trained EBM thus learns to assign low energy values (i.e., high likelihood) to samples drawn from the in-distribution (source)  $Q$, while assigning high energy (i.e., low likelihood) to out-of-distribution samples, such as those from a shifted target distribution $P$, where $P \neq Q$.

\citet{jem} and \citet{yang2021jemimprovedtechniquestraining} present an innovative perspective on reframing discriminative models within the EBM framework.
In their formulation, the energy function for a given input-label pair $(x, y)$ is defined as $E_\phi(x, y) = -f_\phi(x)[y]$, where $f_\phi(x)[y]$ denotes the logit corresponding to label $y$ in the discriminative model $f_\phi$ (i.e., classifier).
Furthermore, the energy function derived from a discriminative model for a single input $x$ can be expressed as the negative \texttt{log-sum-exp} of the logits across all classes in the final classifier layer:
\begin{equation}
  E_\phi(x) = -T \cdot \log \sum_{k=1}^{C} e^{f_\phi(x)[k] / T},
  \label{eqn:eng_fx}
\end{equation}
where $T$ is a temperature parameter that controls the sharpness of the distribution \citep{liu2021energybasedoutofdistributiondetection}. Finally, using a discriminative model within the EBM framework allows one to express the marginal probability of a data sample $x$. 
The main challenge in optimization stems from the normalization constant $Z$, which requires integrating over the entire input space, a task that is generally intractable.
Consequently, EBMs often rely on specialized training methods such as contrastive divergence~\citep{pmlr-vR5-carreira-perpinan05a} or Markov chain Monte Carlo (MCMC) sampling to approximate or avoid the direct computation of $Z$.

% \vspace{-5pt}
\section{Proposed Method: \textsc{CreTTA}}
\label{sec:methods}

% \vspace{-10pt}
\paragraph{Motivation.}
Our focus is on \textit{whether reliable adaptation can be achieved cost-effectively}. 
The reliable adaptation of EBMs has been validated in TEA but this comes with a trade-off of expensive sampling. Instead, AEA~\citep{choiadaptive} directly minimizes target energy within an EM framework without additional sampling. Yet, due to its fundamental reliance on entropy minimization and the lack of proper regularization, it suffers from overfitting, degrading calibration as shown in \autoref{tab:AccCompare}. This outcome is misaligned with the goal of reliable adaptation that motivates our work.

Based on these observations, we question whether the trade-off between reliability and efficiency is truly inevitable. To address this, we propose \textsc{CreTTA}, a sampling-free EBM objective preserves reliable adaptation comparable to TEA, while achieving this at approximately \textit{8× lower computational cost}. Furthermore, we demonstrate that our objective can attain even more stable adaptation than TEA in more dynamic scenarios and on larger-scale datasets.

\paragraph{A Residual Perspective on Distribution Shift.}
\label{subsec:trgdist}

To characterize the distribution shift at test-time, we utilize a \textit{residual energy function} $R$ that captures the discrepancy between the source and target distributions.
Formally, let $Q$ denote the source distribution and $P$ the target distribution. We can express $P$ in terms of $Q$ via an exponential factor encoding the residual energy $R$: $P = Q\exp(-R)/Z$, where $Z$ is a normalization constant. 

By analogy, the marginal distribution of the target data $p_{\theta}$ can be written as the product of the pretrained source model $q_{\phi}$ and an exponential residual term:
\begin{equation}
\resizebox{0.7\linewidth}{!}{$
p_{\theta}(x) = \frac{1}{Z(\theta)}q_{\phi}(x) \exp\left(-\frac{1}{\beta}\tilde{E}_{\theta}(x)\right),
$}
\label{eqn:res_int}
\end{equation}
where the residual energy function $\tilde{E}_{\theta}$ is designed to model only the discrepancy between the fixed source model and the target distribution.
Moreover, $\beta > 0$ is a temperature parameter and $\log Z(\theta)$ is constant across samples.
During TTA, the source model $q_{\phi}$ remains frozen, and $\tilde{E}_{\theta}$ is learned to capture only the distributional differences that arise under domain shift.
In other words, the residual energy function $\tilde{E}_{\theta}$ focuses exclusively on the distribution-shift-induced residuals, refining the energy landscape so that the combined model aligns more closely with the true target distribution while preserving the original source knowledge.

Our residual interpretation of distribution shift can be viewed as an extension of the architectural constraints commonly employed in standard TTA setup. 
Building on the observation that updating only a subset of model parameters, such as the batch-normalization (BN) layers, enables efficient and stable adaptation by mitigating overfitting to severe distribution changes~\citep{wang2021tentfullytesttimeadaptation, wu2024testtimedomainadaptationlearning, zhao2023pitfallstesttimeadaptation}, numerous TTA methods adopt such restricted update strategies.
From a mathematical perspective, we extend this idea by recasting TTA as the problem of learning the residual component of the distribution shift that the pretrained model has not yet captured. This formulation serves as an implicit regularizer: it restricts the target model to \textit{learn only the unmodeled portion of the shift}, thereby limiting the deviation from the source distribution and preventing overfitting, as discussed in \autoref{subsec:performance_comparison}.

\paragraph{Learning Residual Energy via Contrastive Objective.}
\label{subsec:objective}
%normalization constant 제거하는게 중요한 이유
Energy-based models (EBMs) trained with maximum-likelihood estimation (MLE) often suffer from \textit{approximation error} and \textit{prohibitive sampling cost}, primarily due to the need to approximate the intractable partition function~\citep{song2021trainenergybasedmodels}.
These limitations render conventional MLE approaches fundamentally ill-suited for practical TTA, where efficiency and stability are critical.

To overcome these challenges, we propose a \emph{contrastive learning} framework that directly \emph{learns the residual energy function without any estimation of the partition function} $Z$.
Instead of optimizing likelihoods, we operate entirely on pairwise energy differences between source and target samples.
This eliminates the need for sampling from the model distribution, making our method both tractable and scalable for TTA.

Our method shares conceptual similarities with Noise Contrastive Estimation (NCE)~\citep{pmlr-v9-gutmann10a}, in that both use contrastive objectives to bypass normalization.
However, unlike NCE, which retains an implicit dependence on $Z$ through the requirement of globally normalized densities, our formulation dispenses with $Z$ entirely, as it only requires relative energies for learning.

Formally, we reinterpret the residual energy function $\tilde{E}_\theta(x)$ as arising from the density ratio between the target distribution $p_\theta(x)$ and the fixed source model $q_\phi(x)$:
\begin{equation*}
\resizebox{0.65\linewidth}{!}{$
    \tilde{E}_{\theta}(x) = -\beta\left( \log \frac{p_{\theta}(x)}{q_{\phi}(x)} + \log Z (\theta)\right).
    $}
    \label{eqn:RE}
\end{equation*}
Assuming that the residual energy function $\tilde{E}_\theta$ should favor target samples $x_t$ over source samples $x_s$, we  model the probability that the residual energy of a target sample is lower than that of a source sample, as:
%(i.e., assigning lower energy to $x_t$ than to $x_s$), 
\begin{equation*}
\resizebox{0.95\linewidth}{!}{$
\begin{aligned}
P\left(\tilde{E}_{\theta}(x_t) < \tilde{E}_{\theta}(x_s)\right) &= \sigma\left(\tilde{E}_{\theta}(x_s) - \tilde{E}_{\theta}(x_t)\right) \\
&= \sigma\left(\underbrace{\beta \left(\log \frac{p_{\theta}(x_{t})}{q_{\phi}(x_{t})} - \log \frac{p_{\theta}(x_{s})}{q_{\phi}(x_{s})} \right)}_{l(x_s,x_t)}\right),
\end{aligned}$}
\label{eqn:BT}
\end{equation*}
where $\sigma(\cdot)$ denotes the sigmoid function, and the logit term $l(x_s,x_t)$ is defined by using the source energy function $E_\phi$ and the target energy function $E_\theta$ as:
\begin{equation}
\resizebox{0.91\linewidth}{!}{$
l(x_s,x_t)
=\beta (E_\phi(x_t)-E_\phi(x_s))-\beta (E_\theta(x_t)-E_\theta(x_s)).  
$}
\label{eqn:logit}
\end{equation}
During optimization, this objective drives the residual energy function $\tilde{E}_\theta$ to consistently reflect the distribution shift by lowering the relative energy of target samples with respect to source samples, thereby aligning the model with the target distribution while keeping the source model fixed.

Finally, we derive the optimization objective for learning the target model $p_\theta$ with the source model $q_\phi$. 
Given a set of source and target pairs $(x_s, x_t)$, the objective is formulated as minimizing the negative log-likelihood of the probability:
\begin{equation}
\resizebox{0.8\linewidth}{!}{$
\mathcal{L}(\theta;\phi, \mathcal{B}) 
= -\frac{1}{|\mathcal{B}|}\sum_{(x_s,x_t)\sim \mathcal{B}}
  \log \sigma\left(
    l\left(x_s, x_t\right)
  \right),
  $}
\label{eqn:final_obj}
\end{equation}
Crucially, as a consequence of our pairwise contrastive objective, the partition function $Z$ cancels out completely, and no sampling is required unlike in MLE or NCE settings. Instead, we leverage a minimal buffer of source samples $\mathcal{B}_s = \{ x_s^{(i)} \mid i = 1,\dots, |\mathcal{B}_s| \},$ to guide optimization and enable stable contrastive adaptation under test-time distribution shift.
Despite its critical role in stabilizing optimization, the buffer size is negligibly small, imposing virtually no burden in modern memory settings and introducing no practical limitations in real-world deployment.

To optimize our objective (\autoref{eqn:final_obj}), we construct a pairwise mini-batch $\mathcal{B} = \{ (x_s^{(i)}, x_t^{(i)})\}$, where each pair consists of a target sample $x_t^{(i)} \in \mathcal{B}_t$ from the current target stream  and a corresponding source sample $x_s^{(i)} \in \mathcal{B}_s$ randomly drawn from the source buffer $\mathcal{B}_s$. 
We demonstrate that our method maintains consistent performance even when the source buffer size $|\mathcal{B}_s|$ is reduced to as little as $1\%$ of the source dataset, significantly lowering memory overhead.
The robustness of our approach is further validated through an ablation study, as presented in \autoref{tab:buffersize}.

Our proposed objective effectively aligns the model with the target distribution while avoiding the explicit estimation of the residual energy function. 
We reformulate both the source and target models as energy-based models, denoted as $E_\phi$ and $E_\theta$, respectively, following \autoref{eqn:eng_fx}, with the target one initialized as $\theta=\phi$.
This reformulation allows us to express the objective solely in terms of energy functions, eliminating the need for explicit normalization constants through algebraic simplifications.
For a full derivation, we provide the mathematical formulation in \autoref{subsubsecderivations}.

\paragraph{Adaptive Gradient Reweighting.}
The stable adaptation achieved by contrastive residual learning can be clarified through a gradient analysis.
The gradient of our objective in \autoref{eqn:final_obj} is computed as follows:
\begin{equation}
\resizebox{0.9\linewidth}{!}{$
\begin{aligned}
&\nabla_\theta \mathcal{L}(\theta;\phi, \mathcal{B}) 
\\&= 
\frac{1}{|\mathcal{B}|}\sum_{(x_s,x_t)\sim \mathcal{B}}
   % \Bigl[
     w(x_s, x_t) 
     \bigl(
       \beta \nabla_\theta \log p_\theta(x_s)
       - \beta \nabla_\theta \log p_\theta(x_t)
     \bigr)
   % \Bigr]
   , 
\end{aligned}
$}
\label{eqn:objgrad}
\end{equation}
where the gradient terms are adaptively reweighted by the weighting function $w(x_s, x_t)$, which is defined as:
\begin{equation*}
w(x_s, x_t) = \sigma\left(\tilde{E}_{\theta}(x_t) - \tilde{E}_{\theta}(x_s)\right) = \sigma\left(-l\left(x_s, x_t\right)\right).
\end{equation*}

In this context, the term \emph{contrastive} does not merely imply decreasing the energies of target samples or increasing those of source samples. Rather, the gradient weights are modulated by the relative energy levels of paired source-target samples, which promotes stable adaptation (\autoref{subsec:performance_comparison}).
%(see \autoref{subsec:res_overfitting} for further analysis).
If we remove the residual assumption, the pairwise contrastive learning objective reduces to the form in \autoref{eqn:pairwise_non}. In this case, there are no bias terms parameterized by $\phi$, so the gradient magnitude depends solely on the target model’s energies, making the method more prone to overfitting, as illustrated in \autoref{fig:Energy_Trajectories}. A more detailed mathematical discussion of the residual assumption and the pairwise contrastive approach is provided in \autoref{sec:comparative_NCE_pair}.
%\autoref{sec:NCE} and \autoref{sec:PairWise}.

%-------------------------------------------------------------------------
\begin{table*}[t!]
\centering
\caption{{Comparison of mean corruption error (mCE $\downarrow$) and expected calibration error (ECE $\downarrow$) on the CIFAR10-C, CIFAR100-C, and TinyImageNet-C datasets at corruption severity level 5 and the average across severity levels 1-5.} The best results are emphasized in \textbf{BOLD}, while the second-best results are \underline{UNDERLINED}.}
\vspace{-0.5\baselineskip}
\resizebox{0.87\textwidth}{!}{%
\begin{tabular}{@{}lcccccccccccc@{}}
\toprule
              & \multicolumn{4}{c}{\textbf{CIFAR-10-C}}   & \multicolumn{4}{c}{\textbf{CIFAR-100-C}}  & \multicolumn{4}{c}{\textbf{TinyImageNet-C}}             \\ \cmidrule(lr){2-5} \cmidrule(lr){6-9} \cmidrule(lr){10-13} 
            & \multicolumn{2}{c}{\textbf{Severity L5}} & \multicolumn{2}{c}{\textbf{Severity Avg}} & \multicolumn{2}{c}{\textbf{Severity L5}} & \multicolumn{2}{c}{\textbf{Severity Avg}} & \multicolumn{2}{c}{\textbf{Severity L5}} & \multicolumn{2}{c}{\textbf{Severity Avg}} \\ \cmidrule(lr){2-3} \cmidrule(lr){4-5} \cmidrule(lr){6-7} \cmidrule(lr){8-9} \cmidrule(lr){10-11} \cmidrule(lr){12-13} 
\textbf{Method} & \textbf{mCE}    & \textbf{ECE}   & \textbf{mCE}    & \textbf{ECE}   & \textbf{mCE}    & \textbf{ECE}   & \textbf{mCE}    & \textbf{ECE}   & \textbf{mCE}    & \textbf{ECE}   & \textbf{mCE}    & \textbf{ECE}   \\ 
\midrule\midrule
Source  & 100.00\% & 10.18\% & 100.00\% & 5.45\% & 100.00\% & 17.71\% & 100.00\% & 11.73\% & 100.00\% & 16.17\% & 100.00\% & 13.46\% \\ 
\midrule
\multicolumn{13}{l}{\emph{\textbf{Normalization}}}\\
BN Adapt     & 87.44\% & 4.85\% & 101.48\% & 3.15\% & 86.61\% & 8.32\% & 96.81\% & 6.88\% & 94.39\% & \underline{13.66\%} & 98.20\% & \underline{12.12\%} \\ 
\midrule

\multicolumn{13}{l}{\emph{\textbf{Pseudo Labeling}}}\\
PL           & 84.94\% & 10.10\% & 90.03\% & 6.20\% & 93.31\% & 23.81\% & 95.57\% & 16.66\% & 99.36\% & 30.95\% & 98.93\% & 23.47\% \\ 

SHOT         & 73.46\% & 5.42\% & 86.46\% & 3.86\% & \underline{78.71\%} & 8.93\% & \underline{88.48\%} & 7.44\% & 94.01\% & 13.81\% & 97.79\% & 12.24\% \\ 
\midrule

\multicolumn{13}{l}{\emph{\textbf{Entropy Minimization}}}\\
TENT         & 73.91\% & 5.49\% & 86.83\% & 3.89\% & 78.94\% & 8.93\% &88.68\% & 7.47\% & 94.03\% & 13.82\% & 97.81\% & 12.24\% \\ 
ETA         & 87.44\% & 4.85\% & 101.47\% & 3.15\% & 84.44\% & 8.54\% & 94.54\% & 7.10\% & 94.29\% & 13.70\% & 98.08\% & 12.16\% \\ 
EATA         & 87.44\% & 4.85\% & 101.47\% & 3.15\% & 84.38\% & 8.54\% & 94.57\% & 7.11\% & 94.27\% & 13.70\% & 98.07\% & 12.16\% \\ 
SAR          & 81.40\% & 4.79\% & 95.59\% & 3.13\% & 82.43\% & 8.31\% &92.55\% & 6.91\% & 94.31\% & 13.72\% & 98.11\% & 12.16\% \\ 
AEA & \underline{71.69\%} &5.09\% & \underline{85.85\%} &3.73\% &78.79\%  & 9.16\% &88.66\% &7.61\% &93.97\% &13.82\% &97.75\% &12.25\%\\ 
\midrule
\multicolumn{13}{l}{\emph{\textbf{Energy-based Models}}}\\
TEA          & 73.18\% & \textbf{3.83\%} & 88.11\% & \textbf{2.68\%} & 80.47\% & \textbf{7.68\%} & 91.02\% & \textbf{6.33\%} & \underline{93.84\%} & 13.84\% & \underline{97.57\%} & 12.24\% \\ 
% \midrule
\rowcolor{gray!20} 
\textsc{CreTTA}  & \textbf{71.25\%} & \underline{4.15\%} & \textbf{84.40\%} & \underline{2.88\%} & \textbf{78.39}\% & \underline{7.99\%} & \textbf{87.67\%} & \underline{6.82\%} & \textbf{93.20\%} & \textbf{13.52\%} & \textbf{96.31\%} & \textbf{11.85\%} \\ 
\bottomrule
\end{tabular}}
\label{tab:AccCompare}
\vspace{-0.8\baselineskip}
\end{table*}

\section{Experiment}
\label{sec:experiment}
In this section, we present a comprehensive analysis of our proposed method, \textsc{CreTTA}, and conduct a detailed comparison against baselines using various benchmark datasets.

%-------------------------------------------------------------------------
\subsection{Experimental Setup}
\label{subsec:experimental-setup}
\paragraph{Benchmark Datasets and Metrics.}
To evaluate corruption robustness in test-time adaptation, we selected three standard benchmarks: (i) {CIFAR10-C}, (ii) {CIFAR100-C}, and (iii) {TinyImageNet-C}~\citep{hendrycks2019benchmarkingneuralnetworkrobustness}. 
Each dataset contains 15 unique corruption types, categorized into 5 severity levels. In our evaluation, we reported the performance as the average across all corruption types to provide a comprehensive measure and used (i) {mean Corruption Error (mCE) \cite{hendrycks2019benchmarkingneuralnetworkrobustness} and Accuracy (ACC)}, (ii) {Expected Calibration Error (ECE)}, and (iii) {Giga Floating-Point Operations (GFLOPs)} to rigorously evaluate robustness, reliability and practical applicability.

\paragraph{Baselines.}
We compared our method with leading TTA approaches:
% method baselines: (i)~{Source:} The pretrained classifier from the source data that performs inference on test data without adaptation.
(i)~{Normalization-based:} BN-Adapt \citep{schneider2020improving} updates batch normalization statistics for test samples.
(ii)~{Pseudo-labeling-based:} PL \citep{lee2013pseudo} and SHOT \citep{liang2021reallyneedaccesssource} where test samples are filtered based on a confidence threshold, and the model is optimized using these pseudo-labels.
(iii)~{Entropy-based:} TENT \citep{wang2021tentfullytesttimeadaptation}, ETA, EATA \citep{niu2022efficienttesttimemodeladaptation}, SAR \citep{niu2023stabletesttimeadaptationdynamic}, and AEA \citep{choiadaptive} aim to minimize entropy on test samples to align with the target distribution.
(iv)~{Energy-based:} TEA \citep{yuan2024teatesttimeenergyadaptation} adapts to the marginal probability of the target distribution using energy-based learning with SGLD sampling.

\paragraph{Implementation Details.}
In our experiments, we employed WRN-40-2 \citep{zagoruyko2016wide} as backbone for CIFAR10-C and CIFAR100-C, and WRN-28-10 for TinyImageNet-C. Pretrained weights were sourced from RobustBench \citep{croce2021robustbench} and models were trained from scratch if unavailable. We performed online adaptation and evaluation following TENT and TEA. To enhance robustness during adaptation, we incorporated data augmentation into the source buffer, which contained only 10\% of the source dataset. While TEA used a 32×32 resolution, our PACS \citep{li2017deeper} experiment used a higher resolution of 224×224, closer to the original resolution, for larger-scale evaluation.
For details, please refer to appendices. 

%-------------------------------------------------------------------------

\subsection{Performance Comparison}
\label{subsec:performance_comparison}
\paragraph{Classification and Calibration Performance.}

\autoref{tab:AccCompare} reports mean Corruption Error (mCE), focusing on the highest severity level 5 and the average across severity levels (1-5) across all datasets and corruption severities. Our proposed method consistently outperformed all other baselines under corrupted settings, notably achieving the lowest corruption error of 84.40\% on CIFAR10-C. This consistent improvement highlights \textsc{CreTTA}'s adaptability and effectiveness in handling larger and more complex datasets, reinforcing its suitability for real-world test-time adaptation scenarios.

In TTA, model calibration is crucial in real-world applications for quantifying the prediction uncertainty, ensuring reliability under domain shifts and unlabeled data scenarios. We evaluated calibration performance using Expected Calibration Error~(ECE) with 10 bins. While TEA performs well on CIFAR datasets, it fails to maintain the same level of superiority on TinyImageNet-C. In contrast, \textsc{CreTTA} demonstrates strong overall performance across all datasets (\autoref{tab:AccCompare}).

%--------------------------------------------
%imagenet ece acc
\begin{figure}
\centering
 \includegraphics[width=0.65\linewidth]{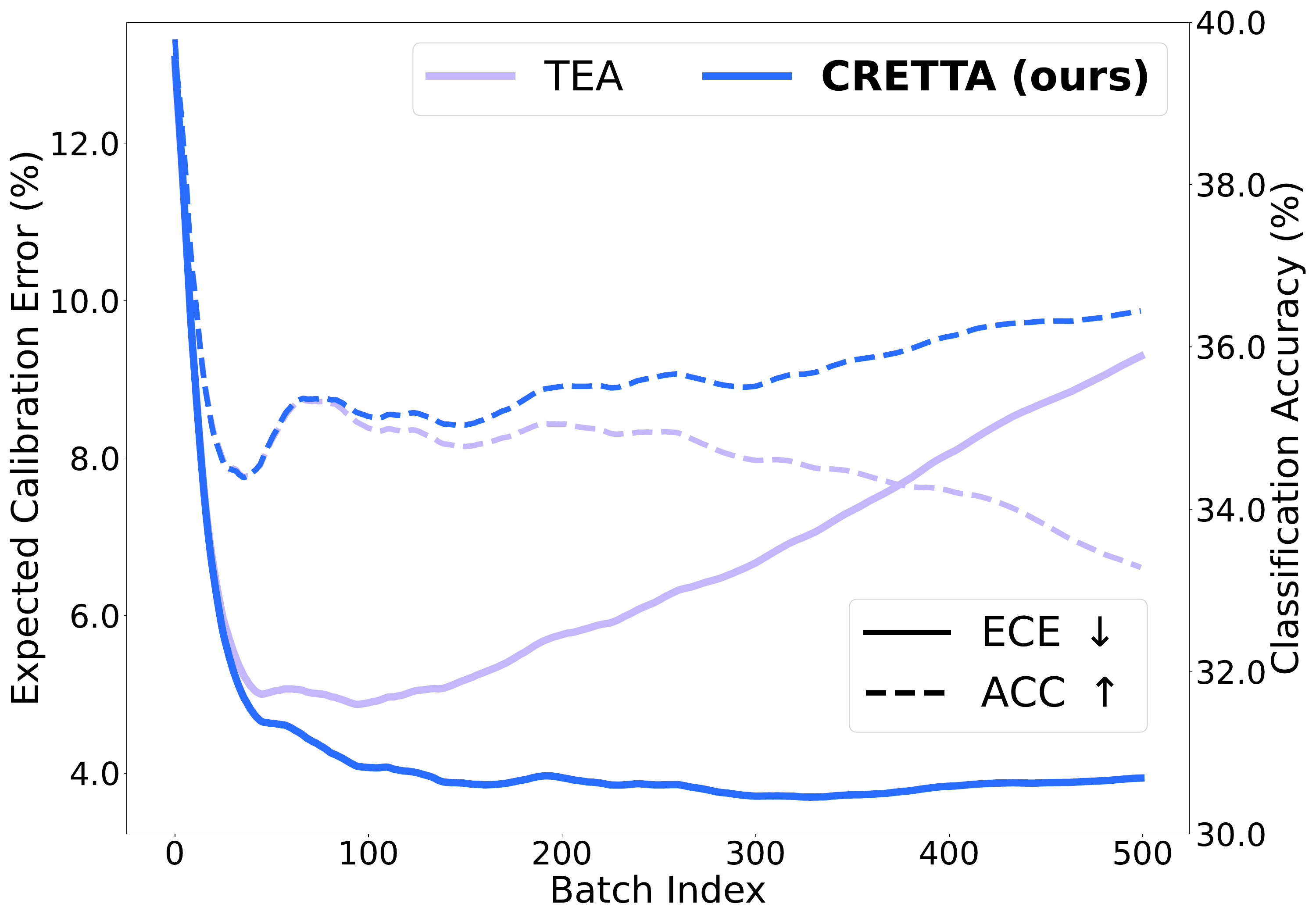}
 \vspace{-0.5\baselineskip}
\caption{ImageNet-C (Sev 5) ECE($\downarrow$) and Acc($\uparrow$) over batch progress. \textsc{CreTTA} maintains stable calibration performance, while TEA experiences approximation error accumulation.}    
\label{fig:imagenetc_ece_acc}
\vspace{-0.5\baselineskip}
\end{figure}
\index{figure}

%--------------------------------------------
%imagenet ece 
\begin{figure}[t]
\centering
\includegraphics[width=0.8\columnwidth]{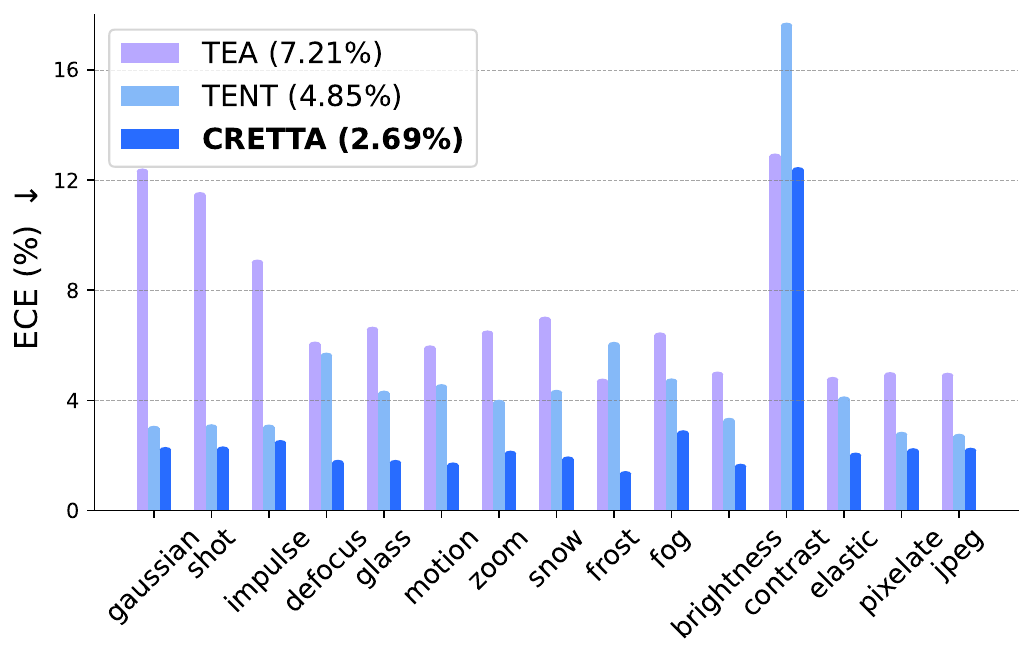}
\vspace{-0.5\baselineskip}
\caption{Comparison of ECE ($\downarrow$) on ImageNet-C across various corruption types, with results averaged over severities 1–5.
} 
\label{fig:ece_image}
\vspace{-0.5\baselineskip}
\end{figure}
\index{figure}

%--------------------------------------------

\paragraph{Scalability.}
Furthermore, we validate that sampling-free design of \textsc{CreTTA} exhibits even more stable adaptation than TEA on large-scale datasets. As the data complexity increases, unresolvable approximation error of TEA becomes pronounced. As shown in \autoref{fig:imagenetc_ece_acc}, TEA manifests a gradual performance degradation both in calibration and accuracy on ImageNet-C. In contrast, \textsc{CreTTA} avoids the approximation procedure and preserves stable improvements throughout the adaptation process. At the end of adaptation, \textsc{CreTTA} achieves substantially better lower calibration error (2.69\%) than TEA (7.21\%), which even underperforms TENT (4.85\%) in \autoref{fig:ece_image}. These trends are consistently observed on another large-scale dataset PACS, which features significant visual domain shifts. In \autoref{tab:pacs_accece}, \textsc{CreTTA} maintains the best classification and calibration performance, outperforming both TENT and TEA by a significant margin.

%--------------------------------------------
%PACS
% \setlength{\intextsep}{6pt}   % 본문-부동체 세로 간격
% \setlength{\columnsep}{12pt}  % 본문-부동체 가로 간격
% \begin{minipage}{\linewidth}

%   % -------- [표1] PACS Accuracy --------
%   \captionof{table}{Acc ($\uparrow$) and ECE ($\downarrow$) on each source domain of PACS.}
%   \label{tab:pacs_accece}  
%   \renewcommand{\arraystretch}{1.0}
%   \centering
%   \begin{adjustbox}{width=1.0\linewidth}
%   \begin{tabular}{llccccc}
%     \toprule
%     & \textbf{Method} & \textbf{P} & \textbf{A} & \textbf{C} & \textbf{S} & \textbf{AVG} \\
%     \midrule\midrule
%     \multirow{3}{*}{Acc} &
%     TENT            & 51.52 & 63.50 & 62.81 & \textbf{37.49} & 53.83 \\
%     &TEA             & 45.55 & 57.95 & 54.24 & 30.80 & 47.14 \\
%     &\textsc{CreTTA} & \textbf{53.06} & \textbf{66.13} & \textbf{64.18} & 36.06 & \textbf{54.86} \\
%     \midrule
%     \multirow{3}{*}{ECE} &
%     TENT            & 44.41 & 35.20 & 34.69 & 58.63 & 43.23 \\
%     &TEA             & 41.71 & 34.62 & 35.02 & \textbf{50.26} & 40.40 \\
%     &\textsc{CreTTA} & \textbf{37.42} & \textbf{28.22} & \textbf{26.65} & 51.68 & \textbf{35.99} \\
%     \bottomrule
%   \end{tabular}
%   \end{adjustbox}
  
% \vspace{-1.0\baselineskip}
% \end{minipage}
% % \vspace{-10pt}

\begin{table}[t]
\setlength{\intextsep}{6pt}
\setlength{\columnsep}{12pt}

\caption{Acc ($\uparrow$) and ECE ($\downarrow$) on each source domain of PACS.}
\vspace{-0.2\baselineskip}
\label{tab:pacs_accece}
\centering
\renewcommand{\arraystretch}{1.0}

\begin{adjustbox}{width=0.8\linewidth}
\begin{tabular}{llccccc}
\toprule
 & \textbf{Method} & \textbf{P} & \textbf{A} & \textbf{C} & \textbf{S} & \textbf{AVG} \\
\midrule\midrule
\multirow{3}{*}{Acc} &
TENT            & 51.52 & 63.50 & 62.81 & \textbf{37.49} & 53.83 \\
& TEA           & 45.55 & 57.95 & 54.24 & 30.80 & 47.14 \\
\rowcolor{gray!20}
\cellcolor{white}
& \textsc{CreTTA} & \textbf{53.06} & \textbf{66.13} & \textbf{64.18} & 36.06 & \textbf{54.86} \\
\midrule
\multirow{3}{*}{ECE} &
TENT            & 44.41 & 35.20 & 34.69 & 58.63 & 43.23 \\
& TEA           & 41.71 & 34.62 & 35.02 & \textbf{50.26} & 40.40 \\
\rowcolor{gray!20}
\cellcolor{white}
 & \textsc{CreTTA} & \textbf{37.42} & \textbf{28.22} & \textbf{26.65} & 51.68 & \textbf{35.99} \\
\bottomrule
\end{tabular}
\end{adjustbox}

\vspace{-0.2\baselineskip}
\end{table}

%--------------------------------------------

%-------------------------------------------------------------------------

%-------------------------------------------------------------------------
\paragraph{Computational Efficiency.}
% A major challenge for previous energy-based TTA methods was the high computational cost for SGLD sampling. This makes them impractical for real-time TTA scenarios that demand rapid adaptation. This computational burden becomes even more pronounced as the input sample size increases. More precisely, TEA not only incurs approximately six times the computational cost (213K GFLOPs) compared to \textsc{CreTTA} (34K GFLOPs) but also struggles to maintain competitive performance. In contrast, \textsc{CreTTA} enables adaptation without explicitly tracking the normalization constant within a pair-wise contrastive learning framework. As shown in \autoref{tab:suppl_comcost_cifar10}, \textsc{CreTTA} consistently outperforms comparison methods, including TENT and BN-Adapt, while maintaining relatively low GFLOPs, demonstrating its efficiency for real-time TTA.

%--------------------------------------------
A major challenge of previous energy-based TTA methods was the high computational cost for SGLD sampling. This makes them impractical for real-time TTA scenarios that demand rapid adaptation. As shown in \autoref{fig:comcost}, TEA incurs an extremely high computational cost of 4335.82 GFLOPs, which is about 32.96$\times$ larger than the source model.

In contrast, \textsc{CreTTA} significantly reduces this overhead, requiring roughly 8$\times$ fewer GFLOPs than TEA, while achieving better classification performance and lower calibration error. This efficiency stems from avoiding explicit estimation of the normalization constant by adopting a pair-wise contrastive learning framework.

Overall, these results demonstrate that \textsc{CreTTA} offers a substantially more scalable and practical solution for real-time TTA than existing energy-based approaches.

% ------------------------------------------------------------
\begin{table}[t!]
\centering
\caption{{Test-time adaptation classification accuracy in dynamic scenarios using CIFAR100-C at severity 5.} Our method demonstrates higher robustness compared to baselines across varying the allocation ratio $\delta$.}
\vspace{-0.5\baselineskip}
\resizebox{0.9\columnwidth}{!}{%
\begin{tabular}{lccccc}
\toprule

\textbf{Method} & $
\boldsymbol{\delta=10}$ & $\boldsymbol{\delta=1}$ & $\boldsymbol{\delta=0.1}$ & $\boldsymbol{\delta=0.01}$ & \textbf{Avg Acc.} \\
\midrule
\midrule
BN Adapt & 61.44\% & 61.11\% & 59.02\% & 45.61\% & 56.79\% \\
\midrule
PL & 44.03\% & 37.27\% & 39.06\% & 43.38\% & 40.93\% \\
SHOT & \underline{63.94\%} & \underline{63.60\%} & \underline{61.20\%} & 46.54\% & 58.82\% \\
\midrule
TENT & 63.91\% & 63.56\% & \underline{61.20\%} & \underline{46.72\%} & \underline{58.85\%} \\
ETA & 62.31\% & 62.04\% & 59.89\% & 46.04\% & 57.57\% \\
EATA & 62.35\% & 62.04\% & 59.84\% & 46.04\% & 57.57\% \\
SAR & 61.54\% & 61.22\% & 59.12\% & 45.66\% & 56.89\% \\
\midrule
TEA & 62.58\% & 62.29\% & 60.08\% & 46.22\% & 57.79\% \\
\midrule
\rowcolor{gray!20} 
\textsc{CreTTA} & \textbf{66.20\%} & \textbf{65.95\%} & \textbf{63.47\%} & \textbf{48.33\%} & \textbf{60.99\%} \\
\bottomrule
\end{tabular}%
}
% }
\label{tab:imbalance}
\vspace{-0.5\baselineskip}
\end{table}

%-------------------------------------------------------------------------
\paragraph{TTA under Non-IID Settings.}
Our previous experiments are conducted under the assumption of i.i.d. test samples which is a widely adopted setting in prior work. Nonetheless, real-world applications can also encounter non-i.i.d. samples \citep{gong2022note, yuan2023robust, wang2022continual}. 
 To further examine the robustness and generalizability of our method beyond the i.i.d. assumption, we constructed a non-i.i.d. test-time adaptation scenario. Specifically, we simulated non i.i.d. data stream by leveraging a Dirichlet distribution to control the class allocation ratio within batch, denoted as $\delta$. A higher $\delta$ value brings the distribution closer to i.i.d., whereas a lower $\delta$ value results in a more non-i.i.d. distribution, where a specific class might dominate the batch. We conducted our experiment on CIFAR100-C using the WRN-28-10 backbone.

As \autoref{tab:imbalance} shows, our method consistently outperforms entropy minimization and instance selection approaches across all $\delta$ values. Specifically, \textsc{CreTTA} achieves the highest average accuracy of 60.99\%,  surpassing TENT’s 58.85\% by 2.14\%p. Also, even at the most imbalanced setting where $\delta = 0.01$, our method achieves a competitive accuracy of 48.33\%. These findings demonstrate that our method not only excels in i.i.d. scenarios but also is effective in dynamic real-world environments.
% ------------------------------------------------------------
\vspace{-5pt}

\begin{table}[t]
\centering
\renewcommand{\arraystretch}{1.1}
\setlength{\tabcolsep}{4pt}

\caption{Classification accuracy under gradual distribution shifts on CIFAR10(-C) and CIFAR100(-C). CRETTA’s residual-energy formulation consistently prevents overfitting and forgetting.
}

\label{tab:gradual}

\resizebox{0.8\columnwidth}{!}{
\begin{tabular}{
>{\centering\arraybackslash}p{1.6cm}
>{\columncolor[gray]{0.9}\centering\arraybackslash}p{1.5cm}
>{\centering\arraybackslash}p{1.3cm}
>{\columncolor[gray]{0.9}\centering\arraybackslash}p{1.5cm}
>{\centering\arraybackslash}p{1.5cm}
}
\toprule
\multirow{2}{*}{\textbf{Domain}} &
\multicolumn{2}{c}{\textbf{CIFAR10}} &
\multicolumn{2}{c}{\textbf{CIFAR100}}\\
\cmidrule(lr){2-3}\cmidrule(lr){4-5}
 & \textbf{\textsc{CreTTA}} & \textbf{TEA} & \textbf{\textsc{CreTTA}} & \textbf{TEA} \\
\midrule\midrule
Source ($Q$) & \textbf{93.46} & 93.45 & \textbf{73.97} & 73.88 \\
1           & \textbf{92.88} & 92.80 & \textbf{71.90} & 71.41 \\
2           & \textbf{92.03} & 91.92 & \textbf{71.57} & 70.40 \\
3           & \textbf{91.63} & 91.29 & \textbf{69.99} & 67.71 \\
4           & \textbf{90.25} & 89.81 & \textbf{67.99} & 65.23 \\
5 ($P$) &
   \textbf{89.47} &
   88.78 &
   \textbf{65.47} &
   60.26 \\
\midrule
Source ($Q$) & \textbf{94.03} & 93.58 & \textbf{75.70} & 69.25 \\
\bottomrule
\end{tabular}
}
\vspace{-\baselineskip}
\end{table}

\vspace{-5pt}

\paragraph{Performance under Gradual Distribution Shift.}
To validate this mechanism, we conducted an experiment under a gradual distribution shift scenario. In this setting, the model continuously adapts from the source distribution $Q$ through increasing shift intensities $(1 \rightarrow 5)$, where severity 5 corresponds to the final target distribution $P$. After the model had fully adapted to $P$, we further froze the model and evaluated its classification accuracy on the original source distribution to observe if the target model remembers the original source distribution. 

As summarized in \autoref{tab:gradual}, \textsc{CreTTA} shows robust adaptation to progressively diverging target distributions, outperforming TEA in classification accuracy on both datasets. Notably, the model’s accuracy on $Q$ improved after adaptation to $P$ (+1.73\%). This provides empirical evidence that \textsc{CreTTA}’s formulation acts as a regularizer that prevents forgetting and facilitates robust adaptation. In contrast, MLE-based method TEA lacks this regularization mechanism and sometimes suffers from forgetting, with performance degradation (-4.63\%) after adaptation.

% \subsection{Resistance Mechanism to Overfitting} 
\subsection{Stability Analysis} 
In this subsection, we further analyze how the residual interpretation on distribution shift and contrastive objective introduced in \autoref{subsec:trgdist} inherently provides stabilized adaptation.

%----------------------------------------------------

\begin{figure}[t]
    \centering
    \includegraphics[width=1.0\linewidth]{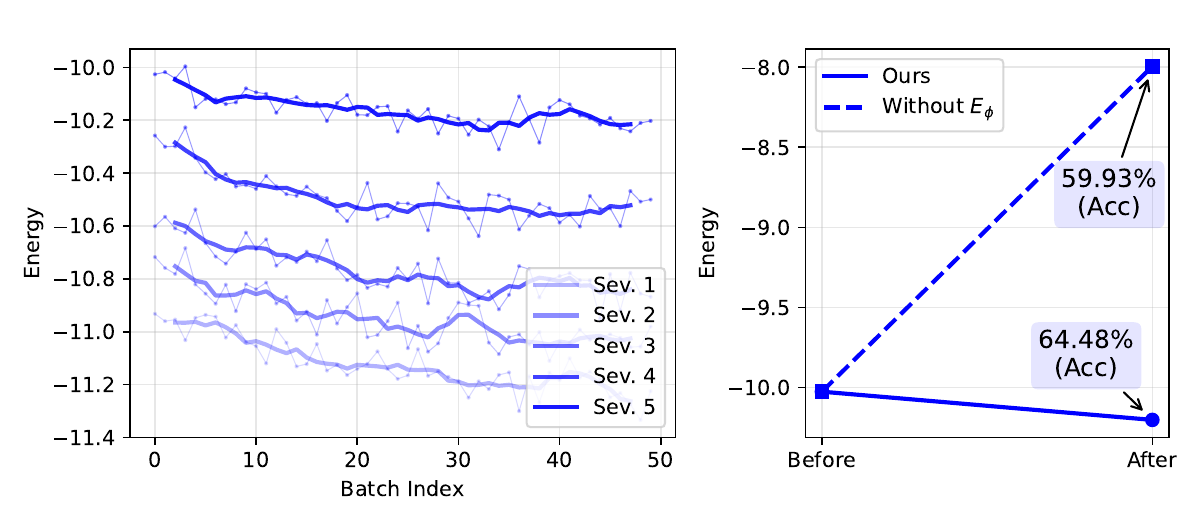}
    \caption{Energy trajectories of target samples across severities (left) and the effect of relative residual energy learning on stable adaptation (right) in CIFAR100-C.}
    \label{fig:Energy_Trajectories}
    \vspace{-\baselineskip}  
\end{figure}

\textsc{CreTTA} tends to reduce target-sample energy during adaptation across severities as shown in \autoref{fig:Energy_Trajectories}, enhancing robustness to strong distribution shifts. While the observed energy reduction is meaningful, the core strengths of \textsc{CreTTA} lie in {contrastive residual energy learning} to prevent convergence to trivial solutions as discussed in \autoref{sec:methods}. The bias terms with respect to $E_\phi$ in \autoref{eqn:logit} prevent $E_\theta$ from becoming excessively small or large, thereby stabilizing the adaptation process. As shown in \autoref{fig:Energy_Trajectories}, the energy of target samples increases drastically after adaptation without bias terms, resulting in performance degradation. 

Consequently, optimization proceeds adaptively based on the relative energy difference between source and target samples. When the energy difference between source and target is already aligned with the desired preference (i.e, $E_\theta(x_t) < E_\theta(x_s)$), the gradient weight $w(x_t, x_s)$ in \autoref{eqn:objgrad} decreases, leading to weaker updates. On the other hand, when the energy difference exists in the opposite direction of the target preference (i.e.,  $E_\theta(x_t) > E_\theta(x_s)$), the gradient weight $w(x_s, x_t)$ increases to enforce stronger corrections. By letting gradient updates depend on these relative energy differences, \textsc{CreTTA} automatically modulates learning strengths through the weighting term, enabling dynamic and stable adaptation.

% %---------------------------------------------------------

%-------------------------------------------------------------------------
\subsection{Ablation Studies}
\paragraph{Contrastive Component Ablation.}
To evaluate the necessity of contrastive components for well-calibrated adaptation, we conduct an objective ablation in \autoref{tab:contrastive_ablation_ece}. Removing the contrastive terms (i.e., $E(x_s)$) and directly minimizing the target energy led to a consistent degradation in calibration across all benchmarks. The effect was particularly pronounced on CIFAR100-C, where the calibration error deteriorated to 11.61\% (+4\%p), which is notably worse than TENT (8.93\%), an entropy minimization--based method that is prone to overconfidence. These results clearly demonstrate that the stability of \textsc{CreTTA}’s adaptation does not arise simply from reducing target energies, but instead benefits from the contrastive learning mechanism.

\begin{table}[h]
\centering
\caption{Ablations on the contrastive component across benchmark datasets with ECE at severity level 5.}
\label{tab:contrastive_ablation_ece}

\label{tab:contrastive_ablation}
\resizebox{0.95\linewidth}{!}{%
\begin{tabular}{rccc}
\toprule
 & \textbf{CIFAR10-C} & \textbf{CIFAR100-C} & \textbf{TinyImageNet-C} \\
\midrule \midrule
w/o Contrastive Terms & 5.57\% & 11.61\% & 16.21\% \\
w/ Contrastive Terms & 4.15\% & 7.99\% & 13.52\% \\
\bottomrule
\end{tabular}
}
% \vspace{-\baselineskip}
\end{table}

% \begin{table}[h]
% \centering
% \caption{Ablations on the contrastive component across benchmark datasets (ECE).}
% \label{tab:contrastive_ablation_ece}

% \label{tab:contrastive_ablation}
% \resizebox{\linewidth}{!}{%
% \begin{tabular}{lcccccc}
% \toprule
%  & CIFAR10-C Sev5 & CIFAR10-C Sev1--5 & CIFAR100-C Sev5 & CIFAR100-C Sev1--5 & TinyIN-C Sev5 & TinyIN-C Sev1--5 \\
% \midrule
% \textbf{W Contrastive Terms (CRETTA)} & 4.15\% & 2.88\% & 7.99\% & 6.82\% & 13.52\% & 11.85\% \\
% \textbf{W/O Contrastive Terms} & 5.57\% & 4.08\% & 11.61\% & 9.65\% & 16.21\% & 14.12\% \\
% \bottomrule
% \end{tabular}
% }
% \end{table}
------------------------------------------------------------
% \begin{wraptable}{r}{0.45\textwidth}
\begin{table}[t]
\centering
\caption{{Effectiveness of buffer size}.}
\resizebox{0.9\columnwidth}{!}{%
\begin{tabular}{cccc}
\toprule
 \textbf{Buffer Ratio} & \textbf{CIFAR10-C} & \textbf{CIFAR100-C} & \textbf{TinyImageNet-C} \\ 
  \midrule \midrule
1\% & 88.00\% & 64.21\% & 40.18\% \\
2\% & 88.17\% & 64.42\% & 40.24\% \\
10\% & 88.30\% & 64.52\% & 40.30\% \\
\bottomrule
\end{tabular}%
}
\label{tab:buffersize}
% \end{wraptable}
% \vspace{-\baselineskip}
\end{table}

% \begin{wraptable}{r}{0.45\textwidth}
\begin{table}[t]
\centering
% \vspace{-5pt}
\caption{{Effectiveness of source buffer content} on CIFAR10-C.}
\resizebox{0.65\columnwidth}{!}{%
\begin{tabular}{ccc}
\toprule
 \textbf{Buffer Type} & \textbf{Sev 5} & \textbf{Sev 1-5 Avg.}  \\ 
  \midrule \midrule
Default(Random) & 88.30\% & 91.01\% \\
CIFAR100-trainset & 88.20\% & 91.02\%  \\
CIFAR100-valset & 88.21\% & 91.03\% \\
\bottomrule
\end{tabular}%
}
% \vspace{-1.0}
\label{tab:bufferunknown}
% \end{wraptable}
\vspace{-\baselineskip}
\end{table}

% \begin{wraptable}{r}{0.5\textwidth}
\begin{table}[t]
\centering
% \vspace{-5pt}
\caption{{Effectiveness of source buffer confidence}.}
\resizebox{0.9\columnwidth}{!}{%
\begin{tabular}{cccc}
\toprule
 \textbf{Buffer Type} & \textbf{CIFAR10-C} & \textbf{CIFAR100-C} & \textbf{TinyImageNet-C} \\ 
  \midrule \midrule
Default (Random) & 88.30\% & 64.52\% & 40.30\% \\
High Confidence & 86.92\% & 64.10\% & 40.18\% \\
Low Confidence & 88.02\% & 64.65\% & 40.92\% \\
\bottomrule
\end{tabular}%
}
% \vspace{-5pt}
% \vspace{-10pt\baselineskip}
\label{tab:bufferconfidence}
% \end{wraptable}
\vspace{-\baselineskip}
\end{table}
% ------------------------------------------------------------
\paragraph{Buffer Ablation.}
Other critical concerns regarding the replay buffer can be summarized in two folds: (i) \emph{Can the source data in the replay buffer be replaced with samples unseen during the pretraining phase?} and (ii) \emph{Does the model maintain its performance regardless of the quality of the samples included in the buffer?}

The first question is critical from a data privacy perspective when the original source data is unavailable. To address this, we evaluate \textsc{CreTTA} on CIFAR10-C with the replay buffer composed of CIFAR100, assumed to be distributionally similar but unseen during pretraining phase. As shown in \autoref{tab:bufferunknown}, no performance drop is observed when the buffer is constructed from either CIFAR100 training or validation set. These findings suggest that \textsc{CreTTA} can operate effectively even without access to the original source data.

We further examined \textsc{CreTTA} in extreme cases where the source buffer composition is biased. Specifically, the buffer was constructed using high confidence (top 10\%) and low confidence (bottom 10\%) samples respectively based on the source model's confidence score. The results summarized in \autoref{tab:bufferconfidence}, indicate that adaptation performance remains unaffected by such variation in buffer content. 

These comprehensive findings highlight that a small and randomly sampled buffer suffices for effective adaptation, regardless of its size or composition. This insensitivity to buffer configuration underscores the practicality of \textsc{CreTTA}, enabling deployment in real-world scenarios with minimal memory overhead and flexible buffer sourcing.

% ------------------------------------------------------------

\section{Related works}
\label{sec: Related works}
\paragraph{Test-time Adaptation.}
TTA is an emerging paradigm that has demonstrated immense potential in adapting pretrained models to unlabeled test data during testing phase. Early methods such as batch normalization adaptations (BN Adapt) \citep{schneider2020improving} leveraged test-batch statistics, while techniques like TTT \citep{sun2020testtimetrainingselfsupervisiongeneralization} and TTT++ \citep{NEURIPS2021_b618c321} utilized image augmentations. TENT \citep{wang2021tentfullytesttimeadaptation}, minimizes entropy to update BN layers, aiming for improved adaptation but often resulting in overconfident that impair model calibration. EATA \citep{niu2022efficienttesttimemodeladaptation} and SAR \citep{niu2023stabletesttimeadaptationdynamic} incorporates instance selection to filter unreliable samples, preserving performance, especially in continual test settings.

\paragraph{Energy-based Models.}
Energy-based models (EBMs) are non-normalized probabilistic models widely used in classification and generative tasks \citep{jem, Guo_2023_ICCV, kim2016deepdirectedgenerativemodels}. 
% JEM \cite{jem} reinterprets discriminative classifiers through the lens of EBMs.
Energy provides a non-probabilistic scalar value capturing the density of the data distribution, making EBMs effective for capturing distribution shifts \citep{NEURIPS2019_378a063b}.
Due to this property, energy-based approaches are utilized in out-of-distribution (OOD) detection and unsupervised domain adaptation \citep{10376546}. Recent works such as AEA \citep{choiadaptive} and TEA \citep{yuan2024teatesttimeenergyadaptation} extend energy to test-time adaptation scenario. 

\paragraph{Learning by Comparison.}
Noise-Contrastive Estimation (NCE) \citep{pmlr-v9-gutmann10a} performs maximum-likelihood estimation through nonlinear logistic regression, distinguishing real data from artificially generated noise. Although it provides insightful ideas as an optimization strategy, NCE still relies on the normalization constant implicitly, which can be difficult to handle in practice. By contrast, pairwise comparison removes the need for this constant by using linear logistic regression. Methods such as RLHF ~\citep{ouyang2022traininglanguagemodelsfollow} and DPO \citep{dpo} use this idea within autoregressive text-generation models to better align generated responses with human preferences. 
%-------------------------------------------------------------------------

\section{Conclusion}
\label{sec:conclusion}

In test-time adaptation, the entropy minimization objective often suffers from poor calibration due to the overconfidence problem, while existing energy-based methods incur significant computational overhead from extensive sampling to approximate the normalization constant for the marginal target distribution. In contrast, \textsc{CreTTA} achieves reliable and efficient adaptation by redefining the distribution shift with a residual energy function while optimizing a contrastive objective that avoids the sampling. 

\textsc{CreTTA} provides two key benefits. First, it inherently mitigates overfitting by adaptively reweighting gradient signals based on relative energy differences, thereby ensuring stable adaptation. Second, by embedding the residual energy function into the contrastive objective, \textsc{CreTTA} eliminates the need for normalization constant approximation. Through comprehensive experiments and ablations \textsc{CreTTA} confirms that it bridges the gap between calibration-aware adaptation and practical feasibility, offering a scalable solution previously unattainable with conventional TTA frameworks.
% -----------------------------------------

% \paragraph{Limitations}
% \textsc{EpoTTA} and many TTA methods \cite{dobler2023robust, sojka2023ar, yu2024stamp, zhou2024resilient} utilize replaying source data or leverage a small buffer to prevent error accumulation and stabilize adaptation process. Despite of their effectiveness, relying on source data and maintaining memory buffer may introduce practical considerations such as limited data accessibility or storage constraints. \textsc{EpoTTA} mitigates these concerns as shown in \autoref{subsec:ablation_study}, by demonstrating that it maintains robust performance even with 1\% of source samples and even with alternative data that is distributionally similar but not part of the original training set. We expect our work to provide valuable implications for applying TTA in practical settings and future works on this end.

\newpage
\section{Impact Statement}
% This paper presents work whose goal is to design an advanced test-time adaptation strategy in real-world scenarios. There are many potential societal consequences of our work, none of which we feel must be specifically highlighted here.

This paper presents work whose goal is to design an advanced test-time adaptation strategy for deployment in real-world scenarios, where models must operate under distribution shifts and limited supervision. By improving the robustness and stability of model adaptation at test time, our approach has the potential to enhance the reliability of machine learning systems in practical applications.

As with many methods that improve model performance and generalization, there may be a range of potential societal consequences associated with this work, depending on the application domain. These consequences could be both beneficial and adverse, and are largely determined by how and where the proposed method is applied. However, we do not identify any specific societal risks or ethical concerns that are unique to our approach or that require separate discussion beyond standard considerations in machine learning research.

\bibliographystyle{icml2026} 
\bibliography{main}

@String(ICLR = {Int. Conf. Learn. Represent.})

@String(ICLR  = {ICLR})

@article{hendrycks2019benchmarkingneuralnetworkrobustness,
  title={Benchmarking neural network robustness to common corruptions and perturbations},
  author={Hendrycks, Dan and Dietterich, Thomas},
  journal={arXiv preprint arXiv:1903.12261},
  year={2019}
}

@inproceedings{sun2020testtimetrainingselfsupervisiongeneralization,
  title={Test-time training with self-supervision for generalization under distribution shifts},
  author={Sun, Yu and Wang, Xiaolong and Liu, Zhuang and Miller, John and Efros, Alexei and Hardt, Moritz},
  booktitle={International conference on machine learning},
  pages={9229--9248},
  year={2020},
  organization={PMLR}
}

@article{NEURIPS2021_b618c321,
  title={Ttt++: When does self-supervised test-time training fail or thrive?},
  author={Liu, Yuejiang and Kothari, Parth and Van Delft, Bastien and Bellot-Gurlet, Baptiste and Mordan, Taylor and Alahi, Alexandre},
  journal={Advances in Neural Information Processing Systems},
  volume={34},
  pages={21808--21820},
  year={2021}
}

@misc{wu2024testtimedomainadaptationlearning,
      title={Test-Time Domain Adaptation by Learning Domain-Aware Batch Normalization}, 
      author={Yanan Wu and Zhixiang Chi and Yang Wang and Konstantinos N. Plataniotis and Songhe Feng},
      year={2024},
      eprint={2312.10165},
      archivePrefix={arXiv},
      primaryClass={cs.CV},
      url={https://arxiv.org/abs/2312.10165}, 
}

@misc{zhao2023pitfallstesttimeadaptation,
      title={On Pitfalls of Test-Time Adaptation}, 
      author={Hao Zhao and Yuejiang Liu and Alexandre Alahi and Tao Lin},
      year={2023},
      eprint={2306.03536},
      archivePrefix={arXiv},
      primaryClass={cs.LG},
      url={https://arxiv.org/abs/2306.03536}, 
}

@misc{press2024entropyenigmasuccessfailure,
      title={The Entropy Enigma: Success and Failure of Entropy Minimization}, 
      author={Ori Press and Ravid Shwartz-Ziv and Yann LeCun and Matthias Bethge},
      year={2024},
      eprint={2405.05012},
      archivePrefix={arXiv},
      primaryClass={cs.CV},
      url={https://arxiv.org/abs/2405.05012}, 
}

@inproceedings{guo2017calibrationmodernneuralnetworks,
  title={On calibration of modern neural networks},
  author={Guo, Chuan and Pleiss, Geoff and Sun, Yu and Weinberger, Kilian Q},
  booktitle={International conference on machine learning},
  pages={1321--1330},
  year={2017},
  organization={PMLR}
}

@article{hendrycks2018baselinedetectingmisclassifiedoutofdistribution,
  title={A baseline for detecting misclassified and out-of-distribution examples in neural networks},
  author={Hendrycks, Dan and Gimpel, Kevin},
  journal={arXiv preprint arXiv:1610.02136},
  year={2016}
}

@article{liu2021energybasedoutofdistributiondetection,
  title={Energy-based out-of-distribution detection},
  author={Liu, Weitang and Wang, Xiaoyun and Owens, John and Li, Yixuan},
  journal={Advances in neural information processing systems},
  volume={33},
  pages={21464--21475},
  year={2020}
}

@article{jem,
  title={Your classifier is secretly an energy based model and you should treat it like one},
  author={Grathwohl, Will and Wang, Kuan-Chieh and Jacobsen, J{\"o}rn-Henrik and Duvenaud, David and Norouzi, Mohammad and Swersky, Kevin},
  journal={arXiv preprint arXiv:1912.03263},
  year={2019}
}

@inproceedings{yang2021jemimprovedtechniquestraining,
  title={Jem++: Improved techniques for training jem},
  author={Yang, Xiulong and Ji, Shihao},
  booktitle={Proceedings of the IEEE/CVF International Conference on Computer Vision},
  pages={6494--6503},
  year={2021}
}

@inproceedings{Guo_2023_ICCV,
  title={EGC: Image Generation and Classification via a Diffusion Energy-Based Model},
  author={Guo, Qiushan and Ma, Chuofan and Jiang, Yi and Yuan, Zehuan and Yu, Yizhou and Luo, Ping},
  booktitle={Proceedings of the IEEE/CVF International Conference on Computer Vision},
  pages={22952--22962},
  year={2023}
}

@article{kim2016deepdirectedgenerativemodels,
  title={Deep directed generative models with energy-based probability estimation},
  author={Kim, Taesup and Bengio, Yoshua},
  journal={arXiv preprint arXiv:1606.03439},
  year={2016}
}

@article{ebmlecun,
  title={A tutorial on energy-based learning},
  author={LeCun, Yann and Chopra, Sumit and Hadsell, Raia and Ranzato, M and Huang, Fujie and others},
  journal={Predicting structured data},
  volume={1},
  number={0},
  year={2006}
}

@article{song2021trainenergybasedmodels,
  title={How to train your energy-based models},
  author={Song, Yang and Kingma, Diederik P},
  journal={arXiv preprint arXiv:2101.03288},
  year={2021}
}

@inproceedings{yair2021contrastivedivergencelearningtime,
  author       = {Omer Yair and
                  Tomer Michaeli},
  title        = {Contrastive Divergence Learning is a Time Reversal Adversarial Game},
  booktitle    = {9th International Conference on Learning Representations, {ICLR} 2021,
                  Virtual Event, Austria, May 3-7, 2021},
  publisher    = {OpenReview.net},
  year         = {2021},
  url          = {https://openreview.net/forum?id=MLSvqIHRidA},
  bibsource    = {dblp computer science bibliography, https://dblp.org}
}

@inproceedings{pmlr-vR5-carreira-perpinan05a,
  title={On contrastive divergence learning},
  author={Carreira-Perpinan, Miguel A and Hinton, Geoffrey},
  booktitle={International workshop on artificial intelligence and statistics},
  pages={33--40},
  year={2005},
  organization={PMLR}
}

@inproceedings{10376546,
  title={Energy-based Self-Training and Normalization for Unsupervised Domain Adaptation},
  author={Herath, Samitha and Fernando, Basura and Abbasnejad, Ehsan and Hayat, Munawar and Khadivi, Shahram and Harandi, Mehrtash and Rezatofighi, Hamid and Haffari, Gholamreza},
  booktitle={Proceedings of the IEEE/CVF International Conference on Computer Vision},
  pages={11653--11662},
  year={2023}
}

@article{NEURIPS2019_378a063b,
  title={Implicit generation and modeling with energy based models},
  author={Du, Yilun and Mordatch, Igor},
  journal={Advances in Neural Information Processing Systems},
  volume={32},
  year={2019}
}

@article{dpo,
  title={Direct preference optimization: Your language model is secretly a reward model},
  author={Rafailov, Rafael and Sharma, Archit and Mitchell, Eric and Manning, Christopher D and Ermon, Stefano and Finn, Chelsea},
  journal={Advances in Neural Information Processing Systems},
  volume={36},
  year={2024}
}

@article{ouyang2022traininglanguagemodelsfollow,
  title={Training language models to follow instructions with human feedback},
  author={Ouyang, Long and Wu, Jeffrey and Jiang, Xu and Almeida, Diogo and Wainwright, Carroll and Mishkin, Pamela and Zhang, Chong and Agarwal, Sandhini and Slama, Katarina and Ray, Alex and others},
  journal={Advances in neural information processing systems},
  volume={35},
  pages={27730--27744},
  year={2022}
}

@article{BT,
  title={Rank analysis of incomplete block designs: I. The method of paired comparisons},
  author={Bradley, Ralph Allan and Terry, Milton E},
  journal={Biometrika},
  volume={39},
  number={3/4},
  pages={324--345},
  year={1952},
  publisher={JSTOR}
}

@inproceedings{niu2022efficienttesttimemodeladaptation,
  title={Efficient test-time model adaptation without forgetting},
  author={Niu, Shuaicheng and Wu, Jiaxiang and Zhang, Yifan and Chen, Yaofo and Zheng, Shijian and Zhao, Peilin and Tan, Mingkui},
  booktitle={International conference on machine learning},
  pages={16888--16905},
  year={2022},
  organization={PMLR}
}

@article{wang2021tentfullytesttimeadaptation,
  title={Tent: Fully test-time adaptation by entropy minimization},
  author={Wang, Dequan and Shelhamer, Evan and Liu, Shaoteng and Olshausen, Bruno and Darrell, Trevor},
  journal={arXiv preprint arXiv:2006.10726},
  year={2020}
}

@article{niu2023stabletesttimeadaptationdynamic,
  title={Towards stable test-time adaptation in dynamic wild world},
  author={Niu, Shuaicheng and Wu, Jiaxiang and Zhang, Yifan and Wen, Zhiquan and Chen, Yaofo and Zhao, Peilin and Tan, Mingkui},
  journal={arXiv preprint arXiv:2302.12400},
  year={2023}
}

@inproceedings{yuan2024teatesttimeenergyadaptation,
  title={TEA: Test-time Energy Adaptation},
  author={Yuan, Yige and Xu, Bingbing and Hou, Liang and Sun, Fei and Shen, Huawei and Cheng, Xueqi},
  booktitle={Proceedings of the IEEE/CVF Conference on Computer Vision and Pattern Recognition},
  pages={23901--23911},
  year={2024}
}

@inproceedings{liang2021reallyneedaccesssource,
  title={Do we really need to access the source data? source hypothesis transfer for unsupervised domain adaptation},
  author={Liang, Jian and Hu, Dapeng and Feng, Jiashi},
  booktitle={International conference on machine learning},
  pages={6028--6039},
  year={2020},
  organization={PMLR}
}

@article{schneider2020improving,
  title={Improving robustness against common corruptions by covariate shift adaptation},
  author={Schneider, Steffen and Rusak, Evgenia and Eck, Luisa and Bringmann, Oliver and Brendel, Wieland and Bethge, Matthias},
  journal={Advances in neural information processing systems},
  volume={33},
  pages={11539--11551},
  year={2020}
}

@inproceedings{lee2013pseudo,
  title={Pseudo-label: The simple and efficient semi-supervised learning method for deep neural networks},
  author={Lee, Dong-Hyun and others},
  booktitle={Workshop on challenges in representation learning, ICML},
  volume={3},
  number={2},
  pages={896},
  year={2013},
  organization={Atlanta}
}

@article{zagoruyko2016wide,
  title={Wide residual networks},
  author={Zagoruyko, Sergey},
  journal={arXiv preprint arXiv:1605.07146},
  year={2016}
}

@article{croce2021robustbench,
  title={Robustbench: a standardized adversarial robustness benchmark},
  author={Croce, Francesco and Andriushchenko, Maksym and Sehwag, Vikash and Debenedetti, Edoardo and Flammarion, Nicolas and Chiang, Mung and Mittal, Prateek and Hein, Matthias},
  journal={arXiv preprint arXiv:2010.09670},
  year={2020}
}

@article{kingma2014adam,
  title={Adam: A method for stochastic optimization},
  author={Kingma, Diederik P},
  journal={arXiv preprint arXiv:1412.6980},
  year={2014}
}

@article{gong2022note,
  title={Note: Robust continual test-time adaptation against temporal correlation},
  author={Gong, Taesik and Jeong, Jongheon and Kim, Taewon and Kim, Yewon and Shin, Jinwoo and Lee, Sung-Ju},
  journal={Advances in Neural Information Processing Systems},
  volume={35},
  pages={27253--27266},
  year={2022}
}

@inproceedings{pmlr-v9-gutmann10a,
  title={Noise-contrastive estimation: A new estimation principle for unnormalized statistical models},
  author={Gutmann, Michael and Hyv{\"a}rinen, Aapo},
  booktitle={Proceedings of the thirteenth international conference on artificial intelligence and statistics},
  pages={297--304},
  year={2010},
  organization={JMLR Workshop and Conference Proceedings}
}

@inproceedings{yuan2023robust,
  title={Robust test-time adaptation in dynamic scenarios},
  author={Yuan, Longhui and Xie, Binhui and Li, Shuang},
  booktitle={Proceedings of the IEEE/CVF Conference on Computer Vision and Pattern Recognition},
  pages={15922--15932},
  year={2023}
}

@inproceedings{wang2022continual,
  title={Continual test-time domain adaptation},
  author={Wang, Qin and Fink, Olga and Van Gool, Luc and Dai, Dengxin},
  booktitle={Proceedings of the IEEE/CVF Conference on Computer Vision and Pattern Recognition},
  pages={7201--7211},
  year={2022}
}

@inproceedings{choiadaptive,
  title={Adaptive Energy Alignment for Accelerating Test-Time Adaptation},
  author={Choi, Wonjeong and Kim, Do-Yeon and Park, Jungwuk and Lee, Jungmoon and Park, Younghyun and Han, Dong-Jun and Moon, Jaekyun},
  booktitle={The Thirteenth International Conference on Learning Representations},
  year={2025}
}

@inproceedings{li2017deeper,
  title={Deeper, broader and artier domain generalization},
  author={Li, Da and Yang, Yongxin and Song, Yi-Zhe and Hospedales, Timothy M},
  booktitle={Proceedings of the IEEE international conference on computer vision},
  pages={5542--5550},
  year={2017}
}

\onecolumn
\appendix
\clearpage
\setcounter{page}{1}
\section{Technical Appendices}

%----------------------------------------------------------------------------
\subsection{Derivation and Function of \textsc{CreTTA}} \label{subsec:derivations_function}
%----------------------------------------------------------------------------
\subsubsection{Derivation of \textsc{CreTTA}} \label{subsubsecderivations}
The marginal distribution of target data $p_{\theta}$ can be written as the product of the pretrained source model $q_{\phi}$ and an exponential residual term:

\begin{equation*}
\label{eqn:residual}
p_{\theta}(x) = \frac{1}{Z}q_{\phi}(x) \exp(-\frac{1}{\beta}\tilde{E}_{\theta}(x)),
\end{equation*}

where $\tilde{E}_{\theta}$ represents the residual energy function encoding the distribution shift.
During TTA, the source model $q_{\phi}$ remains fixed, and the objective is to learn $\tilde{E}_{\theta}$ so as to align the source model more closely with the target distribution. By expanding the equation with respect to the energy function $\tilde{E}_{\theta}$, we can compute the residual energy score of an image sample $x$.

\begin{equation*}
    \tilde{E}_{\theta}(x) = -\beta\left( \log \frac{p_{\theta}(x)}{q_{\phi}(x)} + \log Z\right).
    \label{eqn:RE}
\end{equation*}

Next, we substitute the ground-truth residual energy function $\tilde{E}_{\theta}^*$ into the Bradley-Terry (BT) model \citep{BT}, which only depends on the difference in energy values between source and target pairs:

\begin{equation*}
\begin{aligned}
P(\tilde{E}(x_t) < \tilde{E}(x_s)) &= \frac{1}{1+\exp(-\tilde{E}_{\theta}(x_s) + \tilde{E}_{\theta}(x_t))} = \frac{1}{1+\exp\left(\beta \log \frac{p_{\theta}(x_{s})}{q_{\phi}(x_{s})} - \beta \log \frac{p_{\theta}(x_{t})}{q_{\phi}(x_{t})} \right)},
\end{aligned}
\label{eqn:BT}
\end{equation*}

where $x_t$ and $x_s$ denote the target and source samples, respectively. Here, for pairwise comparison, we use the negative residual energy.

Having derived the probability of the target distribution data in terms of the optimal energy function, which can further be expressed using $\phi$ and $\theta$, our objective for the target model is as follows:

\begin{equation}
\label{eqn:intermediate}
\small
\begin{aligned}
\mathcal{L}(\theta;\phi)
&= -\mathbb{E}_{(x_s,x_t)\sim B} \left[\log \sigma\left(\beta \log \frac{p_{\theta}(x_t)}{q_{\phi}(x_t)} - \beta \log \frac{p_{\theta}(x_s)}{q_{\phi}(x_s)}\right) \right]
\end{aligned}
\end{equation}

In \autoref{sec:methods}, we emphasize that the key advantage of \textsc{CreTTA} is that it avoids the costly Stochastic Gradient Langevin Dynamics (SGLD) sampling required to compute the normalization constant as required in TEA \citep{yuan2024teatesttimeenergyadaptation}. However, the objective \autoref{eqn:intermediate} still includes the normalization constant for both target and source model. 

To eliminate both normalization constants, we first redefine the target and source models using the Gibbs distribution as follows: 

\begin{equation*}
  p_{\theta}(x) = \frac{\text{exp}(-E_{\theta}(x))}{Z({\theta})}\text{ , } q_{\phi}(x) = \frac{\text{exp}(-E_{\phi}(x))}{Z({\phi})}
  % \label{eqn:margdist_tar}
\end{equation*}

By integrating $p_{\theta}$ and $q_{\phi}$ into the \autoref{eqn:intermediate} and applying the logarithm, the normalization constants for both target and source model, i.e., $Z(\theta)$ and $Z(\phi)$, are canceled out as shown in below:

\begin{small}
\begin{align}
\mathcal{L}(\theta;\phi) 
&= -\mathbb{E}_{(x_s,x_t)\sim B} \Big[
\ln \sigma \Big(
\beta \big(-E_{\theta}(x_t) - \cancel{\ln Z(\theta)} + E_{\phi}(x_t) + \cancel{\ln Z(\phi)}\big) \nonumber \\
&\hspace{5em} - \beta \big(-E_{\theta}(x_s) - \cancel{\ln Z(\theta)} + E_{\phi}(x_s) + \cancel{\ln Z(\phi)}\big)
\Big)
\Big]
\label{eqn:cancel}
\end{align}
\end{small}

Therefore, the final learning objective is expressed as follows:

\begin{equation}
\begin{aligned} 
  \mathcal{L}(\theta;\phi) = -\mathbb{E}_{(x_s,x_t)\sim B}\left[\ln \sigma\left(\beta\left(-E_{\theta}(x_t) + E_{\theta}(x_s) + E_{\phi}(x_t) - E_{\phi}(x_s)\right)\right)\right] \nonumber
\end{aligned}
\end{equation}

%----------------------------------------------------------------------------
\subsubsection{Function of \textsc{CreTTA}} \label{subsubsec:lossfunction}
In this section, we provide a detailed explanation of how each component of \textsc{CreTTA} contributes to adaptation, as well as the expected behavior during early and late stages of online adaptation.

If the target model $\theta$ successfully optimizes this objective, then residual energy function $\tilde{E}_{\theta}(x)$ in 
$p_{\theta}(x) = \frac{1}{Z}q_{\phi}(x) \exp(-\frac{1}{\beta}\tilde{E}_{\theta}(x))$
models the residual component of the distribution shift between the source and target domains.

At the beginning of adaptation, the model has not yet encoded the distribution shift. Therefore, the residual energy $E_\theta(x)$ is close to zero for both source samples $x_s$ and target samples $x_t$. 
This results in: $p_\theta(x) \approx q_\phi(x)$
meaning that predictions for both source and target data remain similar to the source model outputs. 

As training progresses, the residual energy function learns the discrepancy between target and source distributions. For source samples $x_s$, $\tilde{E}_\theta(x_s)$ remains small, leading to $p_\theta(x_s) \approx q(x_s)$, preserving source performance. For target samples $x_t$, the residual energy adjusts predictions reflecting the learned domain shift and improving performance on the target domain.
By progressively learning the residual while maintaining alignment with the source model, \textsc{CreTTA} achieves better generalization.

%----------------------------------------------------------------------------

\subsubsection{Buffer Management of \textsc{CreTTA}} \label{subsubsec:buffermanagement}

\textsc{CreTTA} initializes the source buffer $\mathcal{B}_s$ at model initialized, prior to adaptation, by randomly sampling source data up to a fixed buffer size, with an equal number of samples per class. 
During adaptation, the samples in the buffer are used sequentially in batches without any additional sampling or refresh, unlike TEA, thereby incurring no additional computational overhead.

%----------------------------------------------------------------------------

%----------------------------------------------------------------------------

\subsection{Additional Experiments} 
\subsubsection{Detailed Performance Comparison}

\begin{table}[H]
\centering
\caption{Comparison of classification accuracy (Acc $\uparrow$) and expected calibration error (ECE $\downarrow$) on the CIFAR10-C, CIFAR100-C, and TinyImageNet-C datasets at corruption severity level 5, the average across severity levels 1-5, and on clean data. The best adaptation results are emphasized in \textbf{BOLD}, while the second-best results are \underline{UNDERLINED}.}
\resizebox{\textwidth}{!}{%
\begin{tabular}{@{}llccccccccccccccc@{}}
\toprule
            &              & \multicolumn{5}{c}{\textbf{CIFAR-10-C}}                                   & \multicolumn{5}{c}{\textbf{CIFAR-100-C}}                                  & \multicolumn{5}{c}{\textbf{TinyImageNet-C}}                             \\ \cmidrule(lr){3-7} \cmidrule(lr){8-12} \cmidrule(lr){13-17} 
            &              & \textbf{Clean}    & \multicolumn{2}{c}{\textbf{Corr Severity 5}} & \multicolumn{2}{c}{\textbf{Corr Severity 1-5 Avg}} & \textbf{Clean}    & \multicolumn{2}{c}{\textbf{Corr Severity 5}} & \multicolumn{2}{c}{\textbf{Corr Severity 1-5 Avg}} & \textbf{Clean}    & \multicolumn{2}{c}{\textbf{Corr Severity 5}} & \multicolumn{2}{c}{\textbf{Corr Severity 1-5 Avg}} \\ \cmidrule(lr){3-3} \cmidrule(lr){4-5} \cmidrule(lr){6-7} \cmidrule(lr){8-8} \cmidrule(lr){9-10} \cmidrule(lr){11-12} \cmidrule(lr){13-13} \cmidrule(lr){14-15} \cmidrule(lr){16-17} 
& \textbf{Method} & \textbf{Acc($\uparrow$)} & \textbf{Acc($\uparrow$)}    & \textbf{ECE($\downarrow$)}   & \textbf{Acc($\uparrow$)}    & \textbf{ECE($\downarrow$)}   & \textbf{Acc($\uparrow$)} & \textbf{Acc($\uparrow$)}    & \textbf{ECE($\downarrow$)}   & \textbf{Acc($\uparrow$)}    & \textbf{ECE($\downarrow$)}   & \textbf{Acc($\uparrow$)} & \textbf{Acc($\uparrow$)}    & \textbf{ECE($\downarrow$)}   & \textbf{Acc($\uparrow$)}    & \textbf{ECE($\downarrow$)}   \\ 
\midrule\midrule
&Source  & 95.08\% & 81.73\% & 10.18\% & 88.82\% & 5.45\% & 76.28\% & 53.25\% & 17.71\% & 64.11\% & 11.73\% & 59.60\% & 35.12\% & 16.17\% & 43.16\% & 13.46\% \\ 
\midrule
\multicolumn{13}{l}{\emph{\textbf{Normalization}}}\\
& BN Adapt     & 93.59\% & 85.46\% & 4.85\% & 89.12\% & 3.15\% & 72.84\% & 60.74\% & 8.32\% & 65.83\% & 6.88\% & 56.72\% & 39.60\% & \underline{13.66\%} & 44.72\% & \underline{12.12\%} \\ 
\midrule
\multicolumn{13}{l}{\emph{\textbf{Pseudo Labeling}}}\\
    & PL           & \textbf{94.85\%} & 84.85\% & 10.10\% & 90.09\% & 6.20 \% & \textbf{75.98\%} & 56.33\% & 23.81\% & 65.72\% & 16.66\% & \textbf{58.95}\% & 35.40\% & 30.95\% & 43.79\% & 23.47\% \\ 
            & SHOT         & 94.38\% & 87.91\% & 5.42\% & 90.78\% & 3.86 \% & 75.00\% & \underline{64.41\%} & 8.93\% & \underline{68.80\%} & 7.44\% & 56.90\% & 39.84\% & 13.81\% & 44.95\% & 12.24\% \\ 
\midrule
\multicolumn{13}{l}{\emph{\textbf{Entropy Minimization}}}\\
 & TENT         & 94.35\% & 87.84\% & 5.49 \% & 90.74\% & 3.89 \% & 74.95\% & 64.31\% & 8.93\% & 68.73\% & 7.47\% & 56.92\% & 39.83\% & 13.82\% & 44.94\% & 12.24\% \\ 
            & ETA         & 93.72\% & 85.46\% & 4.85 \% & 89.12\% & 3.15 \% & 73.71\% & 61.77\% & 8.54\% & 66.66\% & 7.10\% & 56.82\% & 39.67\% & 13.70\% & 44.79\% & 12.16\% \\ 
            & EATA         & 93.72\% & 85.46\% & 4.85\% & 89.12\% & 3.15\% & 73.66\% & 61.79\% & 8.54\% & 66.65\% & 7.11\% & 56.86\% & 39.68\% & 13.70\% & 44.79\% & 12.16\% \\ 
            & SAR          & 93.61\% & 86.54\% & 4.79\% & 89.80\% & 3.13\% & 73.73\% & 62.71\% & 8.31\% & 67.36\% & 6.91\% & 56.77\% & 39.66\% & 13.72\% & 44.77\% & 12.16\% \\ 
            & AEA & 94.21\% &\underline{88.27\%} &5.09\% & \underline{90.88\%} &3.73\% & 75.17\% &64.40\%  &9.16\% &68.75\% &7.61\% &56.97\% &39.87\% &13.82\% &44.97\% &12.25\%\\
\midrule
\multicolumn{13}{l}{\emph{\textbf{Energy-based Models}}}\\
& TEA          & 94.06\% & 88.06\% & \textbf{3.83\%} & 90.67\% & \textbf{2.68\%} & 74.18\% & 63.66\% & \textbf{7.68\%} & 67.93\% & \textbf{6.33\%} & 57.17\% & \underline{39.96\%} & 13.84\% & \underline{45.08\%} & 12.24\% \\ 
\rowcolor{gray!20} 
            & \textsc{CreTTA}   & \underline{94.43\%} & \textbf{88.30\%} & \underline{4.15\%} & \textbf{91.01\%} & \underline{2.88\%} & \underline{75.26\%} & \textbf{64.52}\% & \underline{7.99\%} & \textbf{69.05\%} & \underline{6.82\%} & \underline{58.23\%} & \textbf{40.30\%} & \textbf{13.52\%} & \textbf{45.75\%} & \textbf{11.85\%} \\ 
\bottomrule
\end{tabular}}
\label{tab:AccCompare_withclean}
\end{table}
%----------------------------------------------------------------------------
\paragraph{Detailed Performance Comparison on Accuracy.}

\autoref{tab:AccCompare_withclean} reports accuracy on the highest severity level 5, the average across severity levels (1-5), and performance on the clean dataset (i.e., without corruption) for CIFAR10-C, CIFAR100-C, and TinyImageNet-C. This table extends the results of \autoref{tab:AccCompare} by additionally reporting accuracy on the clean dataset, providing a more complete view of model performance.

While \textsc{CreTTA} achieves the second-best accuracy on clean data among all methods, with the PL method performing the best. However, this can lead to overfitting and significant degradation in performance under severe corruptions. Notably, while PL exhibits substantial drops in performance under corruption, \textsc{CreTTA} remains robust and effective across both clean and corrupted settings, demonstrating its reliability in both distributions.

%-----------------------------------------------------------------------------------
% \input{table/table2_ece_domain}
%-----------------------------------------------------------------------------------

\begin{figure}[H]
\centering
\includegraphics[width=0.45\columnwidth]{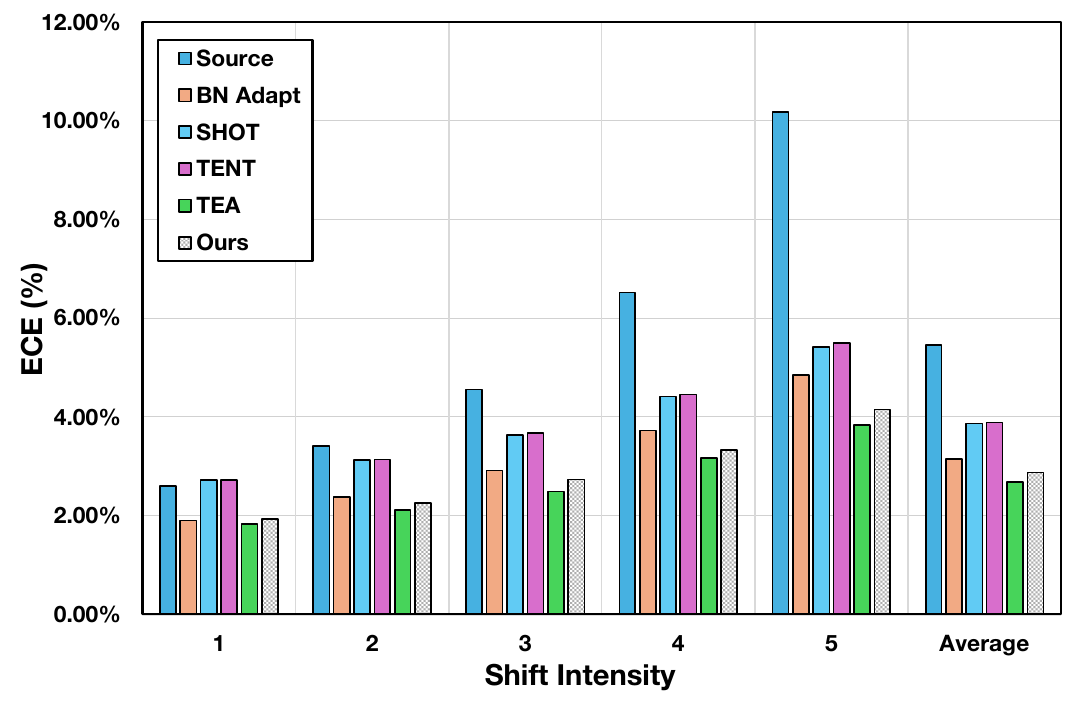}
\caption{Comparison of Expected Calibration Error (ECE$\downarrow$) on the CIFAR10-C dataset across different corruption severity levels.}
\label{fig:ece_cifar10}
\end{figure}
\index{figure}

%-----------------------------------------------------------------------------------

\begin{figure}[H]
\centering
\includegraphics[width=0.45\columnwidth]{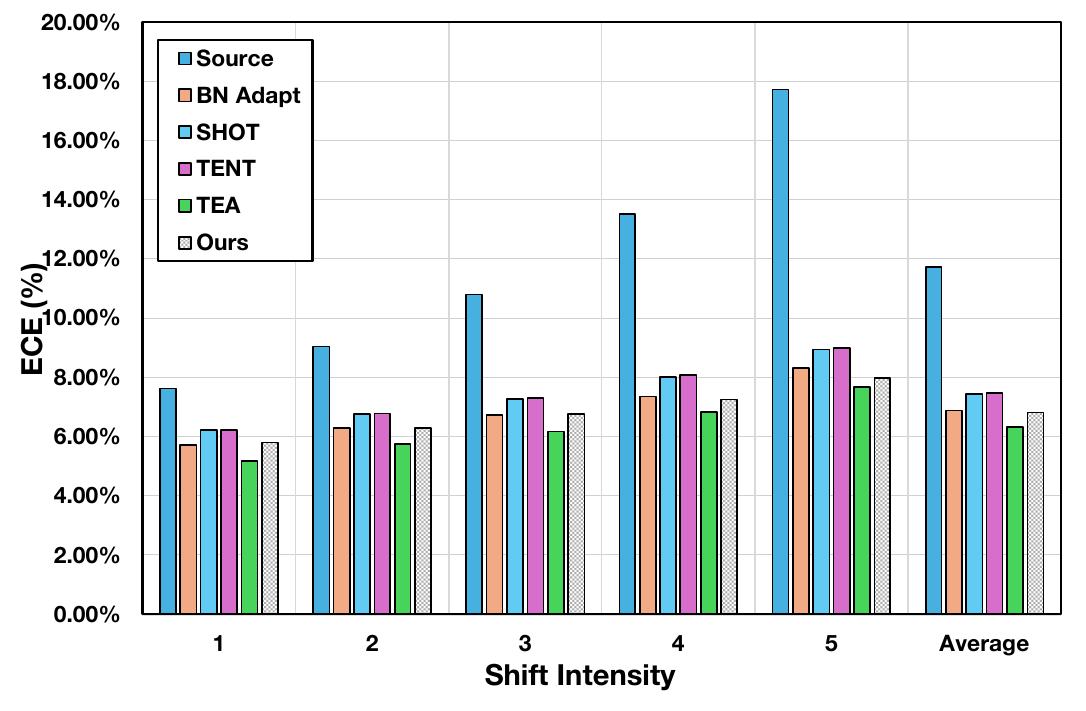}
\caption{Comparison of Expected Calibration Error (ECE$\downarrow$) on the CIFAR100-C dataset across different corruption severity levels.}
\label{fig:ece_cifar100}
\end{figure}
\index{figure}

%-----------------------------------------------------------------------------------

\begin{figure}[H]
\centering
\includegraphics[width=0.45\columnwidth]{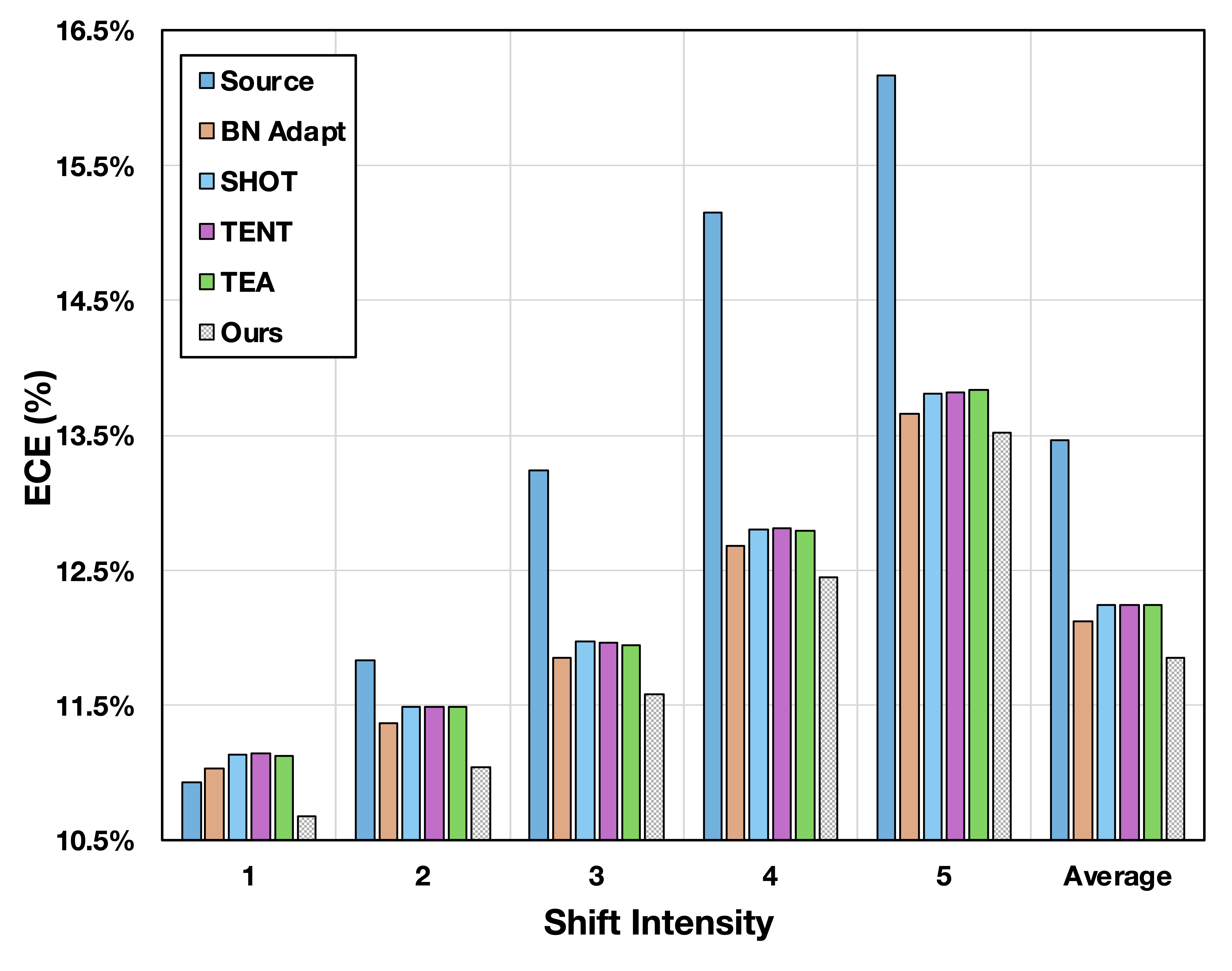}
\caption{Comparison of Expected Calibration Error (ECE$\downarrow$) on the TinyImageNet-C dataset across different corruption severity levels.
} 
\label{fig:ece_imagenet}
\end{figure}
\index{figure}

% \begin{figure}[H]
% \centering
% \includegraphics[width=0.8\columnwidth]{figures/imagenet_ece.pdf}
% \caption{Comparison of Expected Comparison of Expected Calibration Error (ECE $\downarrow$) on ImageNet-C across various corruption types, with results averaged over severities 1–5..
% } 
% \label{fig:ece_image}
% \end{figure}
% \index{figure}

%-----------------------------------------------------------------------------------
\paragraph{Detailed Performance Comparison on Calibration Error.}

In this section, we provide a detailed analysis of the Expected Calibration Error (ECE) for CIFAR10-C and CIFAR100-C. This expands upon the results shown in \autoref{tab:AccCompare}.

As seen in \autoref{fig:ece_cifar10} and \autoref{fig:ece_cifar100}, energy-based methods such as TEA and \textsc{CreTTA} consistently outperform baseline approaches like TENT, which suffers from overconfidence issues. Furthermore, our method maintains computational advantage over TEA, making it more efficient while achieving comparable or superior performance.

On TinyImageNet-C dataset, shown in \autoref{fig:ece_imagenet}, \textsc{CreTTA} outperforms all competing methods across all severity levels. This consistent superiority over all baseline methods demonstrates the robustness and adaptability of our approach in high-complexity datasets. 

%-----------------------------------

\begin{table}[H]
\centering
\caption{Comparison of computational cost (GFLOPs), Memory Cost (Peak Memory Usage), and performance metrics (ECE and Acc) for baselines on the CIFAR10-C}
\resizebox{0.45\linewidth}{!}{
\begin{tabular}{lcccc}
\toprule
 & \textbf{GFLOPs($\downarrow$)} & \textbf{Memory Cost($\downarrow$)} & \textbf{Acc($\uparrow$)} & \textbf{ECE($\downarrow$)} \\
\midrule\midrule
Source      & 131.53  & 443.98 MB  & 88.82\% & 5.45\% \\
BN Adapt    & 131.53  & 452.61 MB & 89.12\% & 3.15\% \\
TENT        & 132.59  & 1546.05 MB & 90.74\% & 3.89\% \\
TEA         & 4335.82 & 3464.78 MB & 90.67\% & 2.68\% \\
\textsc{CreTTA}        & 527.40  & 2651.83 MB & 91.01\% & 2.88\% \\
\bottomrule
\end{tabular}
}
\label{tab:suppl_comcost_cifar10}
\end{table}

% --------------------------------------------------------------------------

\begin{table}[H]
\centering
\caption{Comparison of computational cost (GFLOPs), Memory Cost (Peak Memory Usage), and performance metrics (ECE and Acc) for baselines on the CIFAR100-C}
\resizebox{0.45\linewidth}{!}{
\begin{tabular}{lcccc}
\toprule
& \textbf{GFLOPs($\downarrow$)} & \textbf{Memory Cost($\downarrow$)} & \textbf{Acc($\uparrow$)} & \textbf{ECE($\downarrow$)}  \\
\midrule\midrule
Source      & 131.53  & 443.03 MB & 64.11\% & 11.73\% \\
BN Adapt    & 131.53  & 452.70 MB & 65.83\% & 6.88\% \\
TENT        & 132.59  & 1546.21 MB & 68.73\% & 7.47\% \\
TEA         & 4335.82 & 3465.00 MB & 67.93\% & 6.33\% \\
\textsc{CreTTA}        & 527.40  & 2651.85 MB & 69.05\% & 6.82\% \\
\bottomrule
\end{tabular}
}
\label{tab:suppl_comcost_cifar100}
\end{table}

% -----------------------------------------------

% \paragraph{Detailed Performance Comparison on Computational Efficiency}
% \begin{figure}[t!]
% \centering
% \includegraphics[width=0.45\linewidth]{./figures/tin200_comcost.png}
% \caption{{Comparison of GFLOPs, ECE and Acc} against competitive baselines on TinyImageNet-C at the average across severity levels 1-5.}
% \label{fig:cost}
% \small
% \vspace{-15pt}
% \end{figure}

To further demonstrate the computational advantages of our proposed method, we present a comprehensive comparison of computational cost (GFLOPs), peak GPU memory usage (MB) with performance metrics (Accuracy and ECE) across CIFAR10-C, CIFAR100-C, and TinyImageNet-C, as summarized in \autoref{tab:suppl_comcost_cifar10},\autoref{tab:suppl_comcost_cifar100}.
Compared to TEA, which incurs substantial computational overhead due to SGLD-based sampling, \textsc{CreTTA} reduces GFLOPs by more than eightfold across datasets.
Furthermore, despite incorporating a source buffer, \textsc{CreTTA} maintains a modest peak GPU memory usage, significantly lower than TEA. The peak GPU memory usage is measured as the maximum allocated GPU memory during adaptation. Consequently, \textsc{CreTTA} offers a practical balance between performance, computational cost, and memory efficiency, making it well-suited for deployment in real-world, resource-constrained environments.

% --------------------------------------------------------------------------
% ---acc ece 전체 테이블 버전
\begin{table}[H]
\centering
\small
\caption{Comparison of classification accuracy (Acc $\uparrow$) and expected calibration error (ECE $\downarrow$) on the ImageNet-C dataset.}
\label{tab:imagenet_c}
\resizebox{0.4\textwidth}{!}{%
\begin{tabular}{@{}lcccc@{}}
\toprule
 & \multicolumn{2}{c}{\textbf{Severity L5}} & \multicolumn{2}{c}{\textbf{Severity Avg}} \\
\cmidrule(lr){2-3} \cmidrule(lr){4-5}
\textbf{Method} & \textbf{Acc($\uparrow$)} & \textbf{ECE($\downarrow$)} & \textbf{Acc($\uparrow$)} & \textbf{ECE($\downarrow$)} \\
\midrule\midrule
TENT            & \textbf{37.39\%} & 7.75\% & \textbf{43.78\%} & 4.85\% \\
TEA             & 31.60\% & 8.39\% & 38.72\% & 7.21\% \\
\textsc{CreTTA} & 37.05\% & \textbf{4.43\%} & 43.54\% & \textbf{2.69\%} \\
\bottomrule
\end{tabular}}
\end{table}

\paragraph{Scalability.}
In this section, we provide a detailed results on ImageNet-C. 

As shown in \autoref{tab:imagenet_c}, \textsc{CreTTA} achieves performance by a significant margin, outperforming the entropy-based method TENT and the existing energy-based method TEA in ECE. 

Entropy minimizations's overconfidence and MLE-based approach's approximation error introduced when estimating its normalization constant term leads to poor calibration which is inappropriate in real-wold TTA scenarios. In contrast, \textsc{CreTTA} generalizes well to large-scale datasets, achieving strong predictive performance with superior calibration.

% --------------------------------------------------------------------------
% \begin{wraptable}{r}{}
\begin{table}[H]

\centering
\caption{Comparison of classification accuracy on CIFAR10(-C), CIFAR100(-C) under gradual distribution shift}
\resizebox{0.5\linewidth}{!}{
\begin{tabular}{c|>{\columncolor[gray]{0.9}}ccc|>{\columncolor[gray]{0.9}}ccc}
\toprule
\multicolumn{1}{c}{}& \multicolumn{3}{c}{\textbf{CIFAR10}} & \multicolumn{3}{c}{\textbf{CIFAR100}}\\
\cmidrule(lr){2-4} \cmidrule(lr){5-7}
 \multicolumn{1}{c}{\textbf{Domain}} & \textbf{\textsc{CreTTA}} & \textbf{TEA} & \multicolumn{1}{c}{\textbf{TENT}} & \textbf{\textsc{CreTTA}} & \textbf{TEA} &\textbf{TENT} \\
\midrule\midrule
Source (Q) & \textbf{93.46} & 93.45 & 93.43 &\textbf{73.97} &73.88&73.57 \\
1          & \textbf{92.88} & 92.80 & 92.77 &\textbf{71.90}  &71.41&71.70\\
2          & \textbf{92.03} & 91.92 & 91.92 &\textbf{71.57}  &70.40 &71.36\\
3          & \textbf{91.63 } & 91.29 & 91.35&69.99  &67.71 &\textbf{70.04}\\
4          & \textbf{90.25} & 89.81  & 90.03&67.99 &65.23 &\textbf{68.28} \\
5 (P)      & \textbf{89.47} & 88.78 & 88.58 &\textbf{65.47} &60.26 &65.23 \\
\bottomrule
\end{tabular}
}
\label{tab:gradual_app}
% \end{wraptable}

\end{table}

\paragraph{Detailed Performance Comparison Under Gradual Shift Scenarios.}

In \autoref{subsec:performance_comparison}, we demonstrated that our contrastive residual energy-based learning shows superior performance over CD MLE-based adaptation method TEA. This tendency was consistently observed under the gradual distribution shift setting in \autoref{tab:gradual_app}, and here we additionally report comparisons with TENT. 

For CIFAR10-C, \textsc{CreTTA} maintains the best performance throughtout the shift. 
For CIFAR100-C, \textsc{CreTTA} shows clear gains under stronger shifts. At severity 5, it achieves 65.47\%, notably higher than TEA(60.26\%) and TENT (65.23\%). While TENT is compertitive at mid-level severities, it degrades more under severe shifts.
Overall, \textsc{CreTTA} provides robust adaptation across gradual shifts while preventing forgetting, outperforming both TEA and TENT.

\subsection{Extended Ablation Study and Analysis}
\label{subsec:ablation_study}

% --------------------------------------------------------------------------

\subsubsection{Detailed Ablation study}

% --------------------------------------------------------------------------

\begin{table}[H]
\centering
\caption{Comparison of classification accuracy(Acc) and expected calibration error(ECE) on benchmark datasets between \textsc{CreTTA}(Default) and \textsc{CreTTA} (Loss Term without Source Model) at severity level 5.}
\renewcommand{\arraystretch}{1.1}
\resizebox{0.6\linewidth}{!}{
\begin{tabular}{lcccccc}
\toprule
 &
\multicolumn{2}{c}{\textbf{CIFAR10-C}} &
\multicolumn{2}{c}{\textbf{CIFAR100-C}} &
\multicolumn{2}{c}{\textbf{TinyImageNet-C}} \\
\cmidrule(lr){2-3}\cmidrule(lr){4-5}\cmidrule(l){6-7}
 \textbf{Method}& \textbf{Acc($\uparrow$)} & \textbf{ECE($\downarrow$)} & \textbf{Acc($\uparrow$)} & \textbf{ECE($\downarrow$)} & \textbf{Acc($\uparrow$)} & \textbf{ECE($\downarrow$)} \\
\midrule\midrule
\textsc{CreTTA}    & \textbf{88.30} & \textbf{4.15} & \textbf{64.52} &7.99 & \textbf{40.30} & \textbf{13.52} \\
w/o Source Model Term & 88.09 & 4.66 & 60.02 & \textbf{5.93 }& 37.46 & 14.55 \\
\bottomrule
\end{tabular}
}
\label{tab:loss_abla}
\end{table}

% --------------------------------------------------------------------------

\paragraph{Loss Ablation.}

We observed that eliminating the source model consistently degraded both accuracy and calibration (ECE) in most cases across our benchmark datasets. These results collectively demonstrate that incorporating source model related terms into our contrastive residual learning is essential for stable adaptation.
%---------------------------------------------------------------------------
\paragraph{Contrastive Component Ablation.}

In this section, we clarify why the contrastive component is necessary and how it contributes to stable adaptation.

\begin{itemize}
    \item \textbf{(1) Why --- Contrastive learning is necessary for well-calibrated and stable adaptation.}  
    We demonstrate this by \emph{removing the contrastive term}, showing that direct minimization of target energy leads to unstable energy collapse and degraded calibration across benchmarks.

    \item \textbf{(2) How --- Contrastive learning provides a more informative gradient signal.}  
    We validate this by \emph{replacing source-buffer samples with target samples}, showing that target-only regularization produces weak gradients and yields only marginal adaptation.
\end{itemize}

First, to empirically validate whether contrastive learning is indeed essential for stable adaptation, we first conducted an additional experiment in which we removed the contrastive term and measured the resulting calibration error across the three benchmark datasets.
%---------------------------------------------------------------------------

\begin{table}[h]
\centering
\caption{Ablations on the contrastive component across benchmark datasets with ECE.}
\label{tab:contrastive_ablation_ece_extend}
\resizebox{\linewidth}{!}{%
\begin{tabular}{lcccccc}
\toprule
 & \textbf{CIFAR10-C Sev5} & \textbf{CIFAR10-C Sev1--5} & \textbf{CIFAR100-C Sev5} & \textbf{CIFAR100-C Sev1--5} & \textbf{TinyImageNet-C Sev5} & \textbf{TinyImageNet-C Sev1--5} \\
\midrule \midrule
\textbf{w/ Contrastive Terms (\textsc{CreTTA})} & 4.15\% & 2.88\% & 7.99\% & 6.82\% & 13.52\% & 11.85\% \\
\textbf{w/o Contrastive Terms} & 5.57\% & 4.08\% & 11.61\% & 9.65\% & 16.21\% & 14.12\% \\
\bottomrule
\end{tabular}
}
\end{table}

% \input{table/table_ablation_contrastive}
%---------------------------------------------------------------------------

As shown in \autoref{tab:contrastive_ablation_ece_extend}, removing the contrastive terms (i.e., $E(x_s)$) and directly minimizing the target energy led to a consistent degradation in calibration across all benchmarks. The effect was particularly pronounced on CIFAR100-C, where the calibration error deteriorated to 11.61\% (+4\%p), which is notably worse than TENT (8.93\%), an entropy minimization--based method that is prone to overconfidence. These results clearly demonstrate that the stability of \textsc{CreTTA}’s adaptation does not arise simply from reducing target energies, but instead stems from the contrastive learning mechanism.

To further analyze how the contrastive term contributes to stable adaptation, we also examined the behavior of energy levels and calibration error on CIFAR100-C (Severity 5) during adaptation, comparing \textsc{CreTTA} against the variant where the contrastive terms are ablated.

%---------------------------------------------------------------------------

\begin{table}[h]
\centering
\caption{Target energy and ECE of \textsc{CreTTA} vs.\ without contrastive terms during adaptation on CIFAR100-C (Severity 5).}
\label{tab:contrastive_ablation_energy}
\resizebox{0.6\linewidth}{!}{%
\begin{tabular}{cccccc}
\toprule
 & \multicolumn{2}{c}{\textbf{Target Energy}} & \multicolumn{2}{c}{\textbf{ECE}} \\
\cmidrule(lr){2-3}\cmidrule(lr){4-5}
\textbf{Batch Idx} &
\textbf{w/ Contrastive} &
\textbf{w/o Contrastive} &
\textbf{w/ Contrastive} &
\textbf{w/0 Contrastive} \\
\midrule \midrule
0  & -9.9781  & -9.9781  & 11.11\% & 11.56\% \\
9  & -10.1208 & -10.9536 & 10.48\% & 11.64\% \\
19 & -10.2124 & -11.5247 & 9.48\%  & 12.97\% \\
29 & -10.1896 & -11.7863 & 9.85\%  & 12.35\% \\
39 & -10.1856 & -12.0177 & 8.88\%  & 13.35\% \\
49 & -10.1904 & -12.2531 & 9.15\%  & 12.92\% \\
\midrule 
\textbf{$\Delta$ (Last--First)} & \textbf{-0.21} & \textbf{-2.28} & \textbf{-1.96\%} & \textbf{+1.35\%} \\
\bottomrule
\end{tabular}
}
\end{table}

%---------------------------------------------------------------------------

As shown in \autoref{tab:contrastive_ablation_energy}, we observe that when the contrastive terms are removed and the model directly minimizes the target energy, the energy level drops rapidly during the early stages of adaptation. While this may facilitate fast initial adaptation, it poses a critical risk to stability since aggressively lowering target energies early on sharpens the energy landscape around target samples, which can lead to overfitting. Empirically, we indeed find that removing the contrastive terms results in an overall increase in calibration error.

In contrast, \textsc{CreTTA} reduces the energy level progressively within the contrastive learning framework, enabling a more stable adaptation trajectory. This gradual reduction improves calibration error over time, demonstrating that the contrastive mechanism plays a key role in stabilizing adaptation dynamics.

%---------------------------------------------------------------------------

\begin{table}[h]
\centering
\caption{Gradient coefficient $w$ of \textsc{CreTTA} and Target-as-Source Buffer Data during adaptation on CIFAR100-C (Severity 5).}
\label{tab:trgassrc_ablation_grad}
\resizebox{0.3\linewidth}{!}{%
\begin{tabular}{ccc}
\toprule
\textbf{Batch Idx} & \textbf{\textsc{CreTTA}} & \textbf{W TRG as SRC} \\
\midrule \midrule
0  & 0.490 & 0.304 \\
9  & 0.535 & 0.406 \\
19 & 0.580 & 0.456 \\
29 & 0.580 & 0.465 \\
39 & 0.627 & 0.518 \\
49 & 0.608 & 0.522 \\
\midrule 
\textbf{AVG} & \textbf{0.572} & \textbf{0.454} \\
\bottomrule
\end{tabular}
}
\end{table}

%---------------------------------------------------------------------------

\begin{table}[h]
\centering
\caption{Performance comparison of Target-as-Source buffer setting on CIFAR100-C (Severity 5).}
\label{tab:trgassrc_ablation_accece}
\resizebox{0.4\linewidth}{!}{%
\begin{tabular}{lcc}
\toprule
\textbf{Method} & \textbf{Sev5 Acc} & \textbf{Sev5 ECE} \\
\midrule \midrule
BN Adapt      & 60.74\% & 8.32\% \\
W TRG as SRC  & 61.39\% (+0.65\%p) & 8.19\% \\
\textbf{\textsc{CreTTA}} & \textbf{64.52\% (+3.78\%p)} & \textbf{7.99\%} \\
\bottomrule
\end{tabular}
}
\end{table}

%---------------------------------------------------------------------------

Furthermore, the contrastive learning mechanism proposed in \textsc{CreTTA} that utilizes a small buffer of source (or distributionally similar) samples provides a more meaningful gradient signal than using target samples for regularization. In our previous comment, we illustrated this using an extreme scenario where the source energy, driven by high-energy target samples, becomes sufficiently large that the gradient effectively vanishes. Here, we provide a more realistic explanation focusing on the relative magnitudes of the energy level differences.

Target samples within the same batch are drawn from the same underlying distribution, and so thus it is unlikely for their energy levels to differ significantly. In contrast, source (or distributionally similar) samples originate from distribution that the pretrained model has already learned, making them more likely to exhibit consistently lower energy values than newly encountered target samples. This creates a meaningful energy gap from the target sample energies, which in turn provides a strong gradient signal during adaptation. Consequently, \textsc{CreTTA}’s contrastive learning framework yields notable gains in classification performance while achieving better calibration.

\textsc{CreTTA} provides a substantially more informative learning signal than simply replacing source samples with low-energy target samples. To validate this, we constructed an experimental setup where the 50\% of samples with the lowest model-computed energy values in each target batch served as source-buffer data to compute the source energy $E(x_s)$. We then compared this setup with \textsc{CreTTA}’s adaptation process and performance.

More concretely, we first compared the magnitude of the gradient coefficient $w$ throughout the adaptation process. As shown in \autoref{tab:trgassrc_ablation_grad}, replacing source samples with low energy target samples results in consistently smaller gradient coefficients than \textsc{CreTTA} across the entire adaptation trajectory. This indicates that the model receives weaker learning signals and therefore fails to sufficiently adapt to the target distribution. Consequently, as shown in \autoref{tab:trgassrc_ablation_accece}, the accuracy improvement is only marginal amounting to just +0.65 percentage points compared to BN adapt, which performs adaptation solely through normalization without any learning. In contrast, \textsc{CreTTA} maintains a relatively meaningful gradient coefficient while gradually increasing the learning signal from the early, high-uncertainty stages of adaptation toward later stages. This leads to both improved classification performance and better calibration, ultimately achieving effective and stable adaptation.

Overall, \textsc{CreTTA}’s contrastive learning is an essential component for achieving well-calibrated and stable test-time adaptation. By progressively lowering target energy and thereby reducing calibration error, it enables a stable adaptation process. Moreover, \textsc{CreTTA}’s contrastive learning methodology, which leverages buffer data, is distinctly more effective than approaches that apply only target-sample-based regularization. Notably, \textsc{CreTTA}’s contrastive framework is also robust to variations in buffer-data content and quality, making it highly practical for real-world deployment.

%---------------------------------------------

% \begin{table}[t]
\begin{table}[H]  % ← 페이지 폭의 40%로 설정

\centering
% \vspace{-13pt}
\caption{Effect of Gradient Coefficient}
\renewcommand{\arraystretch}{1.1}
\resizebox{0.4\textwidth}{!}{
\begin{tabular}{lcccccc}
\toprule
 &
\multicolumn{2}{c}{\textbf{CIFAR10-C}} &
\multicolumn{2}{c}{\textbf{CIFAR100-C}} &
\multicolumn{2}{c}{\textbf{TinyImageNet-C}} \\
\cmidrule(lr){2-3}\cmidrule(lr){4-5}\cmidrule(l){6-7}
\textbf{Method} & \textbf{Acc($\uparrow$)} & \textbf{ECE($\downarrow$)}  & \textbf{Acc($\uparrow$)} & \textbf{ECE($\downarrow$)}  & \textbf{Acc($\uparrow$)} & \textbf{ECE($\downarrow$)}  \\
\midrule \midrule
\textsc{CreTTA}    & \textbf{88.30} & 4.15 & \textbf{64.52} & \textbf{7.99} & \textbf{40.30} & \textbf{13.52} \\
Uniform & 87.47 &\textbf{ 4.13} & 61.66 & 8.03 & 38.33 & 15.13 \\
\bottomrule
\end{tabular}
}
\label{tab:grad_coeff_uniform}
\vspace{-0.6\baselineskip}
% \end{wraptable}
\end{table}

%---------------------------------------------------------
\paragraph{Gradient Ablation.}

The gradient coeffcient $w(x_s,x_t)$ is the key mechanism that turns relative energy into stable updates. To verify this role, we conducted an ablation study that disrupts the proposed weighting scheme by replacing $w(x_s,x_t)$ with values randomly sampled from a uniform distribution $[0,1)$. As shown in \autoref{tab:grad_coeff_uniform}, this replacement lead to lower accuracy and higher calibration error, confirming that gradient coefficient is critical for stable optimization and robust adaptation under noisy target data.

% --------------------------------------------------------------------------

\begin{table}[H]
\centering
\caption{Comparison of classification accuracy(Acc) and expected calibration error(ECE) on benchmark datasets between \textsc{CreTTA} (Default) and \textsc{CreTTA} (Single Source Sample in Buffer) at severity level 5.}
\renewcommand{\arraystretch}{1.1}
\resizebox{0.6\textwidth}{!}{
\begin{tabular}{lcccccc}
\toprule
\multirow{2}{*}{\textbf{Method}} &
\multicolumn{2}{c}{\textbf{CIFAR10-C}} &
\multicolumn{2}{c}{\textbf{CIFAR100-C}} &
\multicolumn{2}{c}{\textbf{TinyImageNet-C}} \\
\cmidrule(lr){2-3}\cmidrule(lr){4-5}\cmidrule(l){6-7}
 & \textbf{Acc($\uparrow$)} & \textbf{ECE($\downarrow$)}  & \textbf{Acc($\uparrow$)} & \textbf{ECE($\downarrow$)}  & \textbf{Acc($\uparrow$)} & \textbf{ECE($\downarrow$)}  \\
\midrule \midrule
\textsc{CreTTA}    & \textbf{88.30} & \textbf{4.15} & \textbf{64.52} & \textbf{7.99} & \textbf{40.30} & \textbf{13.52} \\
\textsc{CreTTA} with single source sample & 87.62 & 5.39 & 62.67 & 8.93 & \textbf{40.30} & 14.13 \\
\bottomrule
\end{tabular}
}
\label{tab:buff_abla_single}
\end{table}

% ------------------------------------------------------------
\paragraph{Extended Buffer Ablation.}
While the specific content of the buffer has less impact on performance, as shown in \autoref{tab:buffersize}, this does not imply that the source buffer itself plays a trivial role. To further verify this, we additionally conducted an experiment where the buffer consists of only a single source sample. As shown in \autoref{tab:buff_abla_single}, accuracy dropped by up to 1.7\% and ECE increased by up to 1.2\% across datasets.

To understand why this is the case, recall that source energy is required not merely because of residual learning, but because it is a crucial component of the pairwise contrastive objective. 
Using an arbitrary reference energy as source energy would likely cause training failure. 
From a gradient perspective, \autoref{eqn:objgrad} shows that during early adaptation, a high source energy drives the gradient weight $w$ to collapse to zero, resulting in a trivial solution where learning cannot proceed---mirroring the well-known constraint in Noise Contrastive Estimation (NCE) regarding the choice of the noise distribution. 
To enable effective early adaptation, the buffer must therefore contain samples drawn from a distribution similar to that learned by the pretrained source model, ensuring that these samples receive low energy and provide meaningful contrastive learning signals.

%%내용 추가
\begin{table}[h]
\centering
\caption{Performance Comparison of Source Buffer Contents on CIFAR100-C}
\label{tab:buff_ablation_cifar100}
\resizebox{0.4\textwidth}{!}{

\begin{tabular}{lcc}
\toprule
\textbf{Buffer Type} & \textbf{Severity 5} & \textbf{Severity 1--5} \\
\midrule
\midrule
\textbf{\textsc{CreTTA (Ours)}} & \textbf{64.52\%} & \textbf{69.05\%} \\
CIFAR-10 (train)       & 64.97\% & 69.37\% \\
CIFAR-10 (val)         & 64.95\% & 69.37\% \\
PACS (sketch)          & 63.74\% & 68.25\% \\
\bottomrule
\end{tabular}
}
\end{table}

This behavior is confirmed empirically. 
As shown in \autoref{tab:buff_ablation_cifar100}, \textsc{CreTTA} maintains strong performance on CIFAR100-C when the buffer contains CIFAR-10, which is not original source data but is distributionally similar. In contrast, performance deteriorates when the buffer contains samples with a very different distribution, such as PACS.

Thus, \autoref{tab:bufferunknown} should not be interpreted as suggesting that the absolute source-energy distribution plays a minor role. Rather, it demonstrates that \textsc{CreTTA} remains robust and consistently effective as long as the buffer contains data that are distributionally similar to the source distribution---even without access to the original source dataset. In practice, while true source data may be inaccessible due to privacy constraints, obtaining similar samples is usually far more feasible.

Overall, \textsc{CreTTA}’s superior performance does not arise merely from lowering target energy; instead, it results from the interplay between the residual formulation, 
the pairwise contrastive objective, and their integration, which together provide a stable and powerful adaptation mechanism.

% --------------------------------------------------------------------------

\begin{table}[H]
\centering
\caption{Effectiveness of preference pair size on CIFAR10-C, CIFAR100-C, and TinyImageNet-C.}
\resizebox{0.5\textwidth}{!}{

\begin{tabular}{cccc}
\toprule
\textbf{} & \textbf{CIFAR10-C} & \textbf{CIFAR100-C} & \textbf{TinyImageNet-C} \\ 
\midrule
\midrule
\textsc{CreTTA} w/o CP & 88.30\% & 64.52\% & 40.30\% \\
\textsc{CreTTA} w/ CP & 88.24\% & 64.69\% & 40.44\% \\
\bottomrule
\end{tabular}%
}
\label{tab:grouping}
% \end{wraptable}
\end{table}

% --------------------------------------------------------------------------
\paragraph{Pair Size Ablation.}
\label{subsec:group}
In \textsc{CreTTA}, we assume that the samples in a test batch represent the target distribution, while the source replay buffer represents the source distribution. The loss is computed by forming pairs between target and source samples within each batch, enabling a direct comparison between the two distributions.

To demonstrate the assumption is valid, we examined the impact of increasing the number of pair combinations using a Cartesian Product (CP) to generate all possible combinations of target and source data within each batch. For example, we use 200 pairs for each adaptation in CIFAR10-C, while the Cartesian Product results in 200×200 pairs.

Our results across three datasets summarized in \autoref{tab:grouping} indicate that generating more pairs does not necessarily lead to performance gain. With only a few pairs, \textsc{CreTTA} can efficiently adapt to the target distribution.
% --------------------------------------------------------------------------

\begin{figure}[H]
    \centering
    \includegraphics[width=0.6\linewidth]{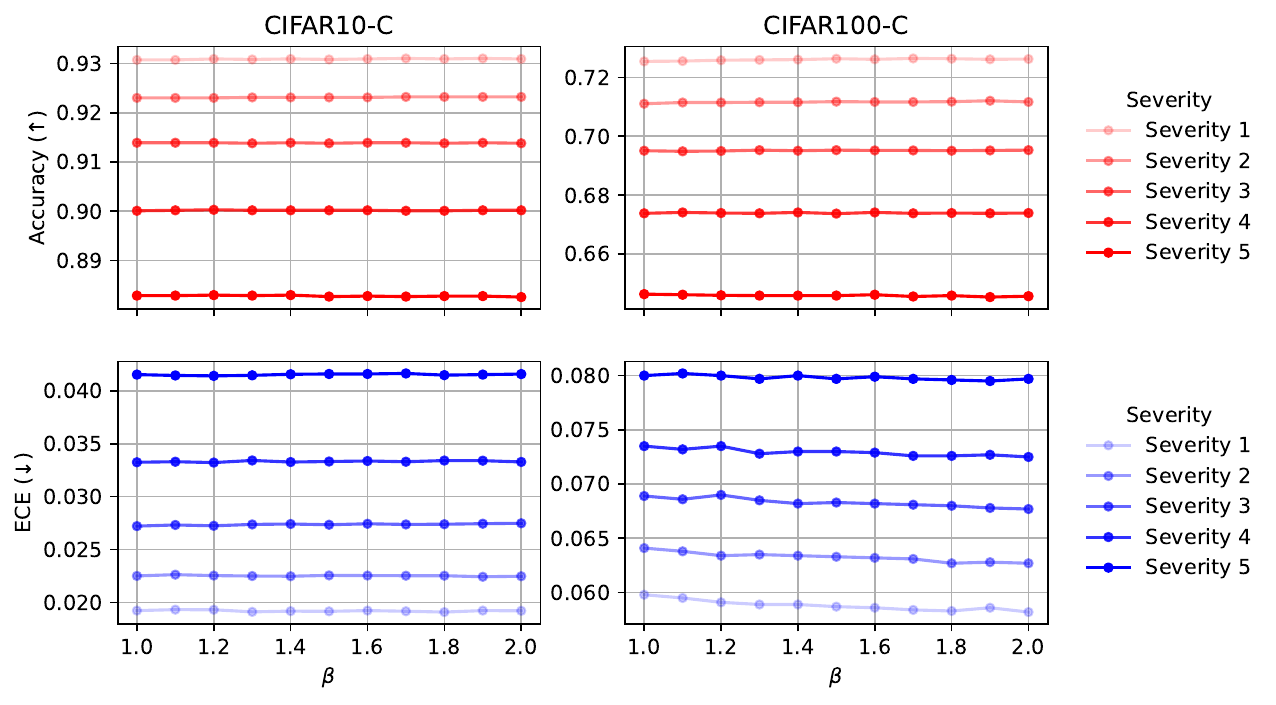}
    \caption{Ablation on varying $\beta$\ values on CIFAR10-C and CIFAR100-C at severity 1-5.}
    \label{fig:beta_ablation_apdx}
\end{figure}

% --------------------------------------------------------------------------
\paragraph{Hyperparameter $\beta$ Ablation.}

\begin{figure}[t]
    \centering
    \includegraphics[width=0.4\textwidth]{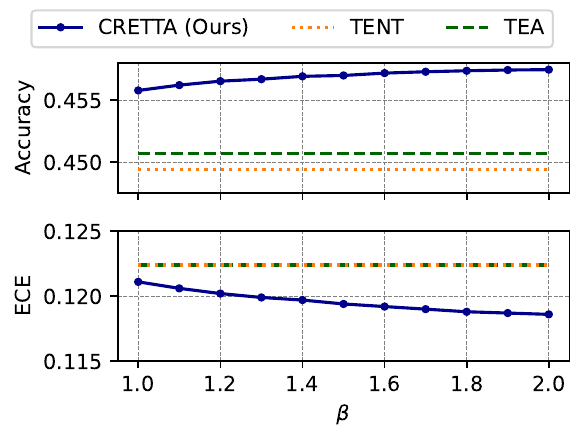}
    \caption{Ablation on varying values of $\beta$.}
    \label{fig:beta}
\end{figure}

The hyperparameter $\beta$ in \autoref{eqn:final_obj} controls the deviation from the pretrained source model, serving as a scaling parameter. To evaluate the robustness of our method, we experiment its performance across varying values of $\beta$, assessing both accuracy and expected calibration error (ECE) on CIFAR10-C, CIFAR100-C and TinyImageNet-C. As shown in \autoref{fig:beta_ablation_apdx}, our method consistently demonstrates stable performance across all corruption severity levels (1-5), validating its robustness. 

In addition, we further examine the effectiveness of \textsc{CreTTA} across varying values of the hyperparameter $\beta$ on TinyImageNet-C, averaging results over severity levels 1 to 5, and compare its performance against competitive baselines (see \autoref{fig:beta}). These results confirm that the strong adaptation performance of \textsc{CreTTA} is not reliant on a specific setting of the temperature parameter $\beta$, but rather stems from our contrastive residual learning objective itself.

% --------------------------------------------------------------------------

\subsubsection{Detailed Setting of \textsc{CreTTA}} \label{subsubsec:detailsetting}

% --------------------------------------------------------------------------

\begin{table}[H]
    \centering
    \caption{Detailed hyperparameters settings for each dataset.}
\resizebox{0.6\textwidth}{!}{
    \begin{tabular}{@{}lcccc@{}}
        \toprule
        \textbf{Dataset} & \textbf{LR} & \textbf{$\beta$} & \textbf{Batch Size} & \textbf{Transformation Type (probability)}\\
        \midrule \midrule
        CIFAR10-C   & 1e-3  & 1.0 & 200 & rotate(1.0)\\
        CIFAR100-C     & 2e-3  & 2.0 & 200 & flip, rotate, affine, perspective, crop(0.2)\\
        TinyImageNet-C & 1e-3  & 2.0 & 1000 & None \\
        \bottomrule
    \end{tabular}
    }
    \label{tab:hyperparam}
\end{table}

% --------------------------------------------------------------------------

\paragraph{Hyperparameters.}
This section details the hyperparameter settings for \textsc{CreTTA}. To optimize performance, minimal hyperparameter tuning was conducted, focusing solely on learning rate, $\beta$ and type and probability of random transformations for source buffer. With only slight adjustments, \textsc{CreTTA} achieved significantly better performance than the current state-of-the-art (SOTA). The batch sizes were aligned with the default settings used in TENT and TEA, which are 200 for CIFAR10-C and CIFAR100-C, 1000 for TinyImageNet-C. For ImageNet-C we follow TENT default settings, using a batch size of 64 and learning rate of $2.5\mathrm{e}{-4}$. We employed the Adam optimizer \citep{kingma2014adam} and reported results over three different random seeds. These settings ensured consistency across experiments while highlighting the robustness and effectiveness of \textsc{CreTTA}. For the PACS domain-generalization task, we used a learning rate of $1\mathrm{e}{-3}$, a batch size of 100, applying source-sample augmentation in the same way as for CIFAR100-C. All experiments were conducted using a single NVIDIA RTX A6000 GPU (48GB).

%-------------------------------------------------------------------------
\paragraph{Evaluation Metrics.}
We evaluate performance on corruption datasets using Average Accuracy and Mean Corruption Error (mCE) following ImageNet-C\citep{hendrycks2019benchmarkingneuralnetworkrobustness}. While Average Accuracy captures absolute classification performance across datasets, Mean Corruption Error (mCE) provides a standardized measure that reflects relative classification error with respect to a base model $f_0$. Following the default evaluation protocol \citep{hendrycks2019benchmarkingneuralnetworkrobustness}, we report mCE in \autoref{tab:AccCompare} to assess classification performance normalized for varying corruption difficulties. The source model without adaptation is used as the base model.

\begin{equation}
\mathrm{mCE}_f=\frac{1}{C} \sum_{c=1}^C \frac{\sum_{s=1}^S \mathcal{E}_{c, s}(f)}{\sum_{s=1}^S \mathcal{E}_{c, s}\left(f_0\right)} 
\end{equation}

Expected Calibration Error (ECE) \citep{guo2017calibrationmodernneuralnetworks} is a metric used to measure the calibration quality of a probabilistic model. Calibration refers to how closely the predicted probabilities of a model match the actual probabilities. ECE quantifies the discrepancy between predicted confidence and actual accuracy.
ECE is calculated as shown in \autoref{eqn:ece}:

\begin{equation}
    \text{ECE} = \sum_{m=1}^{M} \frac{| \text{bin}_m |}{N} \cdot \left| \text{confidence}_m - \text{accuracy}_m \right|    
    \label{eqn:ece}
\end{equation}

where $M$ is the number of bins, $N$ is the total number of data points, $\text{bin}_{m}$ is the number of predictions in $m$-th bin, and $\text{confidence}_{m}$ and $\text{accuracy}_{m}$ are the confidence and accuracy of bin $m$, respectively.

% --------------------------------------------------------------------------

\begin{figure}[h!]
    \centering 
    \includegraphics[width=0.9\textwidth]{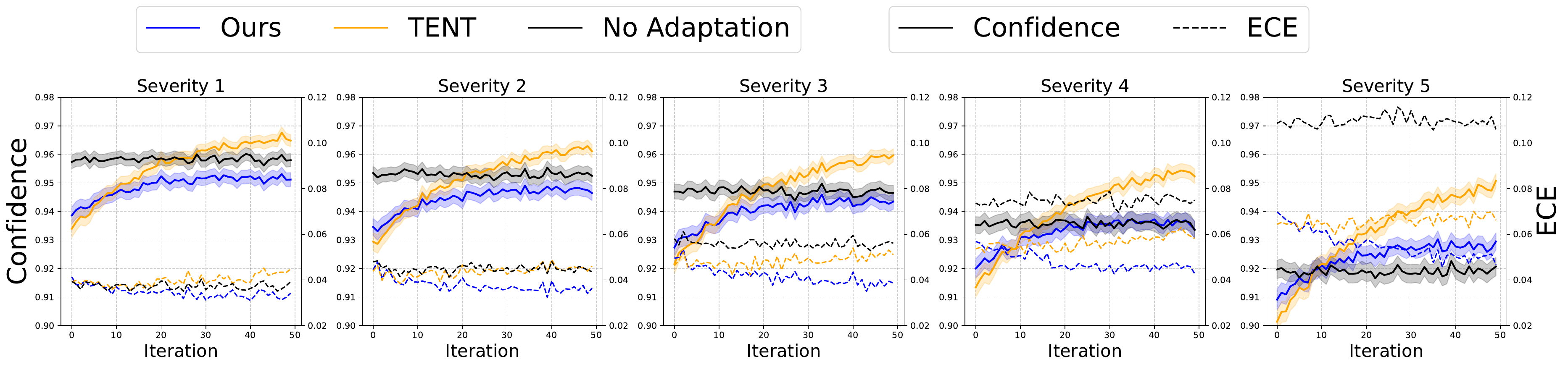}
    % \vspace{-20pt}
    \caption{{The overconfidence problem of entropy minimization in test-time adaptation on CIFAR10-C.} TENT tends to increase a model's confidence in uncertain predictions as adaptation progresses, often leading to worse calibration due to overconfidence. In contrast, \textsc{CreTTA} (Ours) stabilizes the adaptation process by gradually reducing the expected calibration error.}
    \label{fig:overconfidence}
\end{figure}

% --------------------------------------------------------------------------

\begin{figure}[H]
    \centering
    \includegraphics[width=0.9\linewidth]{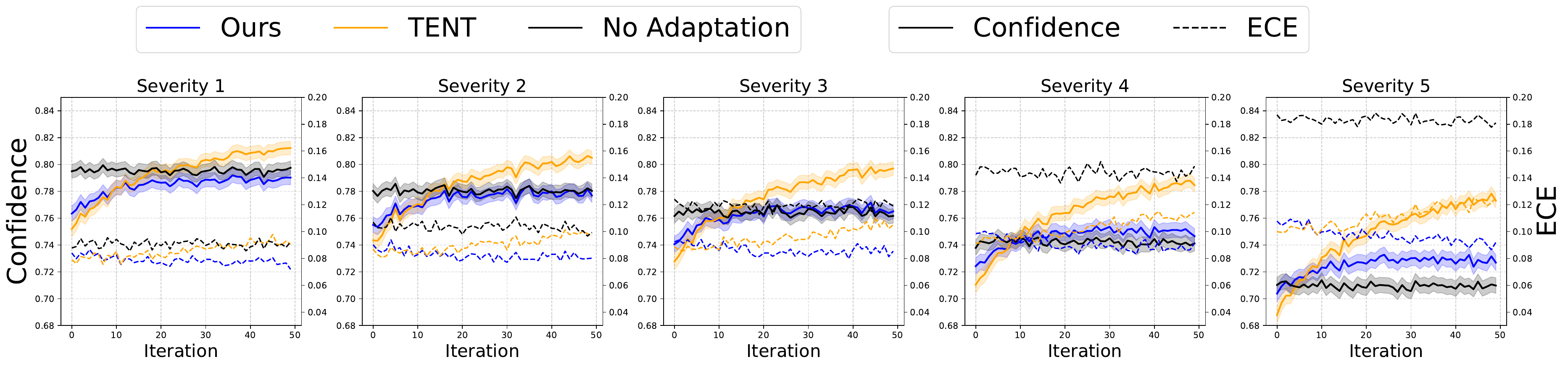}
    \caption{The overconfidence problem of entropy minimization in test-time adaptation on CIFAR100-C.}
    \label{fig:tent_vs_ours_cifar100}
\end{figure}

\begin{figure}[H]
    \centering
    \includegraphics[width=0.9\linewidth]{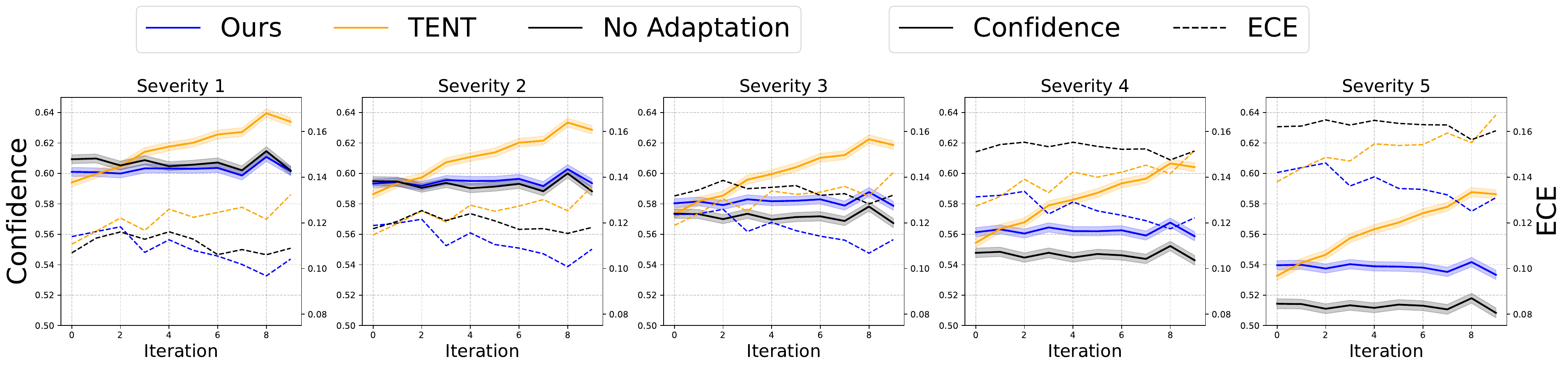}
    \caption{The overconfidence problem of entropy minimization in test-time adaptation on TinyImageNet-C.}
    \label{fig:tent_vs_ours_tin200}
\end{figure}

% --------------------------------------------------------------------------

\subsubsection{Detailed Analysis}
\paragraph{Overconfidence Problem of Entropy Minimization.}
The overconfidence issue inherent in entropy minimization has been thoroughly investigated in prior works~\citep{liu2021energybasedoutofdistributiondetection, hendrycks2018baselinedetectingmisclassifiedoutofdistribution,guo2017calibrationmodernneuralnetworks}. Building on this, we explored that increasing a model's prediction confidence especially when the label information is unavailable can lead to bad calibration as shown in \autoref{fig:ece_cifar10}. The trend consistently appears in other benchmark datasets including CIFAR100-C and TinyImageNet-C as illustrated in \autoref{fig:tent_vs_ours_cifar100} and \autoref{fig:tent_vs_ours_tin200}. Entropy minimization raises the model’s confidence across all severity levels, with the rate of increase becoming steeper as corruption severity intensifies, thereby exacerbating error accumulation.  

On the other hand, \textsc{CreTTA} maintains stable confidence managing uncertainty during test-time adaptation and even reduces calibration error as adaptation progresses. These results suggest that maximizing the marginal likelihood of target samples provides a safer and more effective strategy compared to relying on uncertain predicted probabilities $p_\theta(\hat{y}|x)$ in the test-time learning objective.

%------------------------------------------------------

\paragraph{Clarifying Why Energy Minimization Improves Generalization and How It Differs from Entropy Minimization.}

In this section, we clarify (1) why energy minimization improves generalization capability and (2) how this approach differs, especially from an optimization perspective, from entropy minimization, which is known to suffer from overconfidence issues.

First, to understand why minimizing the energy of target samples leads to improved generalization capability (i.e., higher classification accuracy) on the target distribution, it is essential to examine how optimizing the energy function reshapes the representations. 

We assume a classifier $f_\theta(x) = g(h_{\theta}(x))$ with feature extractor $h_{\theta}$ and a frozen linear classifier $g(z) = Wz + b$, where $W = [w_1, \dots, w_C]^\top \in \mathbb{R}^{C \times d}$. The logits can be expressed as
\[
a_k(x) = w_k^\top h_\theta(x) + b_k.
\]

Since target samples contain no labels, we optimize the marginal energy of the input,
\[
E_\theta(x) = -\log \sum_{k=1}^C \exp(a_k(x)),
\]
which is the standard unnormalized negative log-density in energy-based models. The gradient of this energy with respect to the logit $a_k(x)$ is
\[
\frac{\partial E_\theta(x)}{\partial a_k(x)} = - \frac{\exp(a_k(x))}{\sum_{k'=1}^C \exp(a_{k'}(x))} = - p_\theta(y=k \mid x).
\]
Using $\frac{\partial a_k}{\partial z} = w_k$, the gradient of the energy with respect to the feature representation becomes
\[
\frac{\partial E_\theta(x)}{\partial z} = \sum_{k=1}^C \frac{\partial E_\theta(x)}{\partial a_k} \frac{\partial a_k}{\partial z}
= - \sum_{k=1}^C p_\theta(y=k \mid x) w_k
= - \mathbb{E}_{p_\theta(y \mid x)}[w_y].
\]

This expression shows that the gradient descent updates shift the feature $z$ toward the expected classifier weight vector. Although the classifier remains frozen, this shift alters the logits and thus reshapes the conditional distribution, enabling the model to align its predictions to the target domain solely through feature-level adjustments. From a representation perspective, energy minimization adapts the feature extractor to better capture the target-domain distribution, producing stronger representations that improve classification accuracy under distribution shift even without labels.

Furthermore, to understand how energy minimization and entropy minimization objectives behave differently during optimization, it is essential to compare their gradients. Since we already expressed the gradient of the energy objective, we now present the gradient expressions for the entropy objective.

For unlabeled target data, the entropy of model prediction is expressed as
\[
\mathcal{L}_{\mathrm{ent}}(x) = - \sum_{k=1}^C p_k \log p_k,
\]
where $p_k = p_\theta(y=k \mid x)$. The corresponding gradient is
\[
\frac{\partial \mathcal{L}_{\mathrm{ent}}}{\partial z}
= - \sum_{y=1}^C (\log p_y + 1)p_y \left( w_y - \sum_{k=1}^C p_k w_k \right).
\]

Letting $\mathbb{E}[w] = \sum_k p_k w_k$ denote the expectation of classifier weights under the current predictive distribution, we obtain the final compact form:
\[
\frac{\partial \mathcal{L}_{\mathrm{ent}}}{\partial z}
= - \sum_{y=1}^C (\log p_y + 1)p_y \left( w_y - \mathbb{E}[w] \right).
\]

In the gradient, the term $(\log p_y + 1)p_y$ heavily weights classes for which $p_y$ is already large and $\log p_y$ is less negative. Thus, the gradient of entropy moves the feature $z$ in the direction that increases confidence for the most likely classes, effectively reducing prediction entropy and behaving similarly to pseudo-labeling. By directly modifying the conditional distribution, entropy minimization mainly pushes the model to become more confident, often excessively.

In contrast, the gradient of the energy objective depends only on $\mathbb{E}_{p_\theta(y\mid x)}[w_y]$, which is a smooth expectation over classifier weights. It does not contain the entropy term’s confidence-amplifying multiplier. This trend is also empirically confirmed in \autoref{fig:overconfidence}, \autoref{fig:tent_vs_ours_cifar100}, and \autoref{fig:tent_vs_ours_tin200}.

Overall, energy minimization improves generalization by altering the logits and reshaping the conditional distribution, allowing the model to align its predictions to the target domain. While achieving strong classification performance, it is also more robust than entropy minimization because it optimizes a smooth expectation over classifier weights, avoiding the entropy objective’s confidence-amplifying multiplier. Together, these properties make energy-based adaptation a more balanced and principled optimization approach, enhancing representation quality rather than merely increasing confidence, which explains its superior robustness in test-time adaptation.

% \begin{algorithm}[H]
% \caption{Contrastive Residual Energy Test-Time Adaptation}

% \textbf{Require:} Source buffer 
% $\mathcal{B}_s = \{ x_s^{(i)} \}_{i=1}^{|\mathcal{B}_s|}$

% \textbf{Require:} Pretrained source model $p_{\phi}$

% \textbf{Require:} Adaptation model $p_{\theta}$ initialized as $p_{\phi}$

% \For{For each adaptation step from $n = 1$ to $T$,

% \begin{algorithmic}[1] 
    
%     \State Observe target mini-batch $\mathcal{B}_t^{(n)}$

%     \State Sample a pairwise mini-batch 
%     $\mathcal{B}^{(n)} = \{(x_s^{(j)}, x_t^{(j)})\}_{j=1}^{|\mathcal{B}^{(n)}|}$, where:
%     \begin{itemize}
%         \item $x_t^{(j)} \in \mathcal{B}_t^{(n)}$ is a target sample from the current stream
%         \item $x_s^{(j)} \in \mathcal{B}_s$ is sampled uniformly from the source buffer $\mathcal{B}_s$
%     \end{itemize}

%     \State Compute loss:
%     \[
%     \mathcal{L}(\theta; \phi, \mathcal{B}) 
%     \quad \text{(see Eq.~\ref{eqn:final_obj})}
%     \]

%     \State Compute gradient:
%     \[
%     g_t = \nabla_{\theta} \mathcal{L}(\theta; \phi, \mathcal{B})
%     \]

%     \State Update parameters:
%     \[
%     \theta \leftarrow \theta - \gamma g_t
%     \]
% \end{algorithmic}
% }
% \textbf{Return:} Adapted model $p_{\theta_T}$
% \end{algorithm}

\section{Comparative Analysis of Noise Contrastive and Pair-wise Contrastive Estimation}
\label{sec:comparative_NCE_pair}
%--------------------------------------------------------------------

\subsection{Noise Contrastive Estimation}
\label{sec:NCE}

We first define a reward function $r(\cdot)$ to properly compare samples from two different sets or distributions. 
\begin{equation*}
r(x;\theta,\phi) = \log P_\theta(x) - \log P_\phi(x)
\end{equation*}
where $P_\theta$ is the target distribution and $P_\phi$ is the source distribution.

%--------------------------------------------------------------------

\subsubsection{Non-residual}
If we define energy functions for each of them by utilizing gibbs distribution,
\begin{equation*}
E_\theta(x) = -\log P_\theta(x) - \log Z(\theta)
\end{equation*}
\begin{equation*}
E_\phi(x) = -\log P_\phi(x) -\log Z(\phi)
\end{equation*}
Then the reward function becomes
\begin{equation*}
r(x;\theta,\phi) = - (E_\theta(x) - E_\phi(x)) + C
\end{equation*}

Then the loss function becomes

\begin{align*}
\mathcal{L}(\theta;\phi) &= -\mathbb{E}_{x_t}\left[\log \sigma (r(x;\theta,\phi))\right]-\mathbb{E}_{x_s}\left[\log (1 -\sigma (r(x;\theta,\phi)))\right] \\
\end{align*}

%--------------------------------------------------------------------

\subsubsection{Residual}
If we define a residual energy function,
\begin{equation*}
p_\theta(x) = \frac{1}{Z}q_\phi(x)\text{exp}(-\frac{1}{\beta} \tilde{E}_\theta(x))
\end{equation*}
Then the reward function becomes
\begin{equation*}
r(x;\theta,\phi) = \log p_\theta(x) - \log q_\phi(x) = -\frac{1}{\beta} \tilde{E}_\theta(x) + c
\end{equation*}
Then the loss function becomes

\begin{align*}
\mathcal{L}(\theta;\phi) &= -\mathbb{E}_{x_t}\left[\log \sigma (r(x;\theta,\phi))\right]-\mathbb{E}_{x_s}\left[\log (1 -\sigma (r(x;\theta,\phi)))\right] \\
\end{align*}

%--------------------------------------------------------------------

\subsection{Pair-wise Contrastive Estimation}
\label{sec:PairWise}
We first define a reward function $r(\cdot)$ to properly compare samples from two different sets or distributions. 
\begin{equation*}
r(x_t, x_s) = \tilde{r}(x_t) - \tilde{r}(x_s)
\end{equation*}
where $\tilde{r}$ is a reward function that assigns higher values to target samples than source samples

%--------------------------------------------------------------------

\subsubsection{Non-residual}
If we define energy functions for each of them,
\begin{equation*}
E_\theta(x) = -\log P_\theta(x) -\log Z(\theta)
\end{equation*}
Then the reward function becomes
\begin{equation*}
r(x_t, x_s) = \log P_\theta(x_t) - \log P_\theta(x_s) = - (E_\theta(x_t) - E_\theta(x_s))
\end{equation*}
Then the loss function becomes
\begin{align*}
\mathcal{L}(\theta;\phi) &= -\mathbb{E}_{x_t, x_s}\left[\log \sigma (r(x_t, x_s))\right]\\
&= -\mathbb{E}_{x_t, x_s}\left[\log \sigma (- (E_\theta(x_t) - E_\theta(x_s)))\right]
\end{align*}
The gradient becomes
\begin{align*}
\nabla_\theta\mathcal{L}(\theta;\phi) &= -\mathbb{E}_{x_t, x_s}\left[\sigma (-r(x_t, x_s))\nabla_\theta r(x_t, x_s)\right]\\
&= \mathbb{E}_{x_t}\left[\sigma (E_\theta(x_t) - E_\theta(x_s)) \left(\nabla_\theta E_\theta(x_t) - \nabla_\theta E_\theta(x_s) \right)\right]\\
\end{align*}
\label{eqn:pairwise_non}

%--------------------------------------------------------------------

\section{Use of LLMs}
We used a large language model (ChatGPT) only as a general purpose assistive tool for minor editing tasks such as polishing sentences, correcting grammar and spelling and making small LaTeX table formatting adjustments. The LLM was not involved in research ideation, experimental design, data analysis, or substantive writing. All technical decisions, interpretations, and the writing of the core content were carried out entirely by the authors, who take full responsibility for the originality of the manuscript.

%--------------------------------------------------------------------

\end{document}